\documentclass[10pt,twocolumn,letterpaper]{article}
\usepackage[T1]{fontenc}
\usepackage[utf8]{inputenc}
\usepackage{times}
\usepackage{iftex}
\ifPDFTeX\pdfminorversion=7\fi
\usepackage[margin=0.75in,top=0.8in,bottom=0.8in,columnsep=0.25in]{geometry}
\usepackage[hyphens]{url}
\usepackage{graphicx}
\usepackage{natbib}
\let\cite\citep
\setcitestyle{aysep={}}
\usepackage{bibunits}
\usepackage{caption}
\usepackage{algorithm}
\usepackage{algorithmic}
\usepackage{booktabs,multirow,tabularx,array}
\usepackage{placeins,flafter}
\usepackage{amsmath,amssymb}
\usepackage[hidelinks]{hyperref}
\DeclareCaptionStyle{ruled}{labelfont=normalfont,labelsep=colon,strut=off}
\graphicspath{{paper/figures/}}
\defaultbibliographystyle{pafs-author-year}
\makeatletter
\newcommand{\pafsbibcontext}[1]{%
  \def\@extra@b@citeb{#1}%
  \def\@extra@binfo{#1}}
\makeatother
\hypersetup{pdftitle={PAFS: Perturbation-Aware Firing Scaffolds for Robust Event-Based SNNs},
 pdfauthor={Ning Liu, Jun Sun, Hao Guo, Sen Shen, Xiangfei Yang, Jie Yang, Lan Wu, Shangke Lyu, Thomas Bräunl}}
\title{PAFS: Perturbation-Aware Firing Scaffolds for Robust Event-Based SNNs}
\author{\begin{minipage}{\dimexpr\textwidth-2\tabcolsep\relax}
\centering
\normalsize
Ning Liu\textsuperscript{1}\quad Jun Sun\textsuperscript{5}\quad Hao Guo\textsuperscript{5}\quad
Sen Shen\textsuperscript{2}\quad Xiangfei Yang\textsuperscript{3,*}\\[4pt]
Jie Yang\textsuperscript{4}\quad Lan Wu\textsuperscript{1}\quad
Shangke Lyu\textsuperscript{5,*}\quad Thomas Br\"aunl\textsuperscript{1}\\[8pt]
\footnotesize
\textsuperscript{1}The University of Western Australia, Australia\\[2pt]
\textsuperscript{2}The Chinese University of Hong Kong, Hong Kong SAR, China\\[2pt]
\textsuperscript{3}Hangzhou City University, China\qquad
\textsuperscript{4}Westlake University, China\qquad
\textsuperscript{5}Nanjing University, China
\end{minipage}}
\date{}
\begin{document}
\maketitle
\begin{abstract}
Event-based spiking neural networks (SNNs) encode information through temporal firing dynamics, but the same stateful computation makes their predictions sensitive to event loss, timestamp jitter, and speed variation. These perturbations alter evidence accumulation, yet late-layer firing trajectories can retain error-diagnostic structure that is absent from the final logits alone. We introduce Perturbation-Aware Firing Scaffolds (PAFS), a labeled post-training calibration method that exploits this structure without updating the pretrained backbone. PAFS associates standardized late-layer leaky integrate-and-fire trajectories with the class-probability residuals of an Adapter anchor, then applies singular value decomposition to extract compact residual-aligned channel directions. At inference, firing trajectories from the same backbone pass are projected onto this fixed low-rank scaffold and combined with anchor statistics to predict a residual logit correction. Across four matched dataset--backbone settings spanning three event datasets and two spiking Transformer architectures, PAFS improves corrupted-average and worst-condition accuracy over the corresponding Adapter anchors, with maximum gains of 3.06 and 3.65 percentage points. In the DailyAction-DVS/Spikingformer setting, the added trainable parameters amount to approximately 0.34\% of the frozen backbone.
\end{abstract}
\begin{bibunit}[pafs-author-year]
\pafsbibcontext{.main}
\section{Introduction}

Event-based spiking neural networks (SNNs) are well suited to asynchronous sensory streams because they combine sparse spike-based computation with stateful temporal processing~\cite{maass1997networks,lichtsteiner2008dvs}. Their membrane states and spike sequences preserve how evidence evolves over time, while recent spiking Transformers further strengthen representation capacity through attention-based feature interaction~\cite{zhou2023spikformer,zhou2026spikingformer}. This combination of sparse computation and temporal state makes SNNs particularly attractive for event-based recognition, where predictions depend not only on which events occur, but also on when and in what order they arrive.

The same stateful dynamics, however, also create a source of vulnerability~\cite{luo2026space}. In event-based systems, sensor noise, bandwidth constraints, synchronization errors, varying observation windows, and changes in motion speed can lead to event loss, timestamp jitter, and temporal rescaling. Although these distortions often preserve the underlying semantics, they alter when and how evidence is accumulated, thereby perturbing membrane dynamics and increasing the risk of incorrect predictions~\cite{hao2024threaten}. This vulnerability is particularly consequential in latency-sensitive applications such as autonomous driving and unmanned aerial vehicle navigation, where perception failures can compromise safe decision-making. Yet the resulting firing responses are not necessarily unstructured: late-layer trajectories may still retain systematic signals associated with the anchor model's residual errors. This motivates a complementary view of temporal robustness: it should not only reduce sensitivity to temporal perturbations, but also exploit the diagnostic information preserved in internal firing dynamics.

Existing SNN robustness methods typically rely on robust training, adversarial optimization, or redesigned neuronal dynamics and event representations~\cite{liang2022toward,ding2022snnrat,ding2024enhancing,xu2024feelsnn}. These approaches require access to the training pipeline and backbone parameters, which may be unavailable when a pretrained model must be adapted to deployment shifts. Parameter efficient methods such as Adapters and LoRA reduce adaptation cost~\cite{houlsby2019parameter,hu2022lora}, but prediction-loss-driven adaptation does not explicitly identify which internal temporal responses remain informative about the model's unresolved errors. We hypothesize that these residual errors are associated with a compact set of coordinated directions in late-layer firing trajectories, providing a targeted signal for labeled post-training calibration with a frozen backbone.

Based on this hypothesis, we propose Perturbation-Aware Firing Scaffolds (PAFS), a lightweight robustness-calibration framework for frozen SNNs. PAFS first trains a bottleneck Adapter on labeled clean and corrupted training samples to obtain an anchor predictor. It then associates standardized late-layer leaky integrate-and-fire (LIF) trajectories with the anchor's class-probability residuals and extracts the dominant residual-aligned channel directions. During inference, a zero-initialized linear head maps the projected scaffold features and anchor statistics to a residual logit correction. PAFS therefore reuses internal temporal information from the same backbone pass; it neither updates the pretrained backbone nor requires a second backbone evaluation.

Across four matched dataset--backbone settings spanning three event-based benchmarks and two spiking Transformer architectures, PAFS improves both corrupted-average and worst-condition accuracy over the corresponding Adapter anchor, while extensive ablations validate the contributions of residual alignment, low-rank scaffold construction, and complementary late-layer firing features.

Our contributions are threefold:

\begin{itemize}

\item We identify that residual prediction errors under temporal
corruptions remain systematically encoded in late-layer firing
trajectories of frozen event-based SNNs, providing correction-relevant
information beyond the final logits.

\item We propose PAFS to operationalize this mechanism by aligning
standardized LIF trajectories with class-wise probability residuals
and extracting a compact residual-aligned firing subspace for
single-pass logit correction.

\item Experiments across three event datasets and two spiking
Transformer backbones demonstrate consistent robustness gains.
Controlled comparisons further show that the benefit arises from
residual-aligned internal firing structure, rather than logits alone,
generic low-rank compression, or a single firing source.

\end{itemize}

\section{Related Work}

\subsection{Event Based Neuromorphic Vision}

Event cameras asynchronously report changes in log intensity as events carrying spatial location, timestamp, and polarity~\cite{lichtsteiner2008dvs}. Their low latency and high dynamic range support fast motion perception and sparse sensing~\cite{gallego2022event}. DVS-Gesture established a widely used event based gesture benchmark~\cite{amir2017low}, while DailyAction-DVS introduces more varied action dynamics~\cite{liu2021event}. DVS-Gesture-Chain further shows that recognition can depend on temporal composition rather than event counts alone. These properties make missing windows, timestamp jitter, and speed changes meaningful robustness tests.

\subsection{Deep SNN Training and Spiking Transformers}

The non differentiable spike function motivates spatio temporal backpropagation~\cite{wu2018stbp} and surrogate gradient learning~\cite{neftci2019surrogate}. Threshold dependent normalization improves optimization in deep SNNs~\cite{zheng2021going}, while residual connections enable greater depth~\cite{fang2021deep}. Temporal efficient training reduces the cost of learning across time~\cite{deng2022temporal}, and learnable LIF dynamics improve temporal representation~\cite{fang2021plif}. Gated attention coding introduces adaptive temporal control~\cite{qiu2024gated}, whereas two compartment neurons enrich state dynamics~\cite{zhang2024tclif}. These methods strengthen the backbone but do not identify which late firing trajectories predict residual errors after training.

Spikformer introduces spiking self attention for event recognition~\cite{zhou2023spikformer}. Spike driven Transformer reduces multiplication heavy computation~\cite{yao2023spikedriven}, and its second version develops a broader meta architecture~\cite{yao2024spikedriventransformerv2}. QKFormer studies hierarchical spiking attention~\cite{zhou2024qkformer}. Spikingformer further uses spike driven residual learning to reduce non spike computation~\cite{zhou2026spikingformer}. PAFS operates on late LIF responses and is evaluated with both Spikformer and Spikingformer.

\subsection{Robustness of SNNs}

Early analysis examined whether spike based models possess inherent robustness~\cite{sharmin2020inherent}. Training based defenses subsequently introduced adversarial objectives~\cite{liang2022toward}, gradient regularization~\cite{ding2022snnrat}, and sparse gradient constraints~\cite{liu2024sparse}. Stochastic gating provides another route to robust spike propagation~\cite{ding2024enhancing}. Event frequency encoding instead modifies the input representation~\cite{xu2024feelsnn}, while membrane potential distribution regularization targets the internal state~\cite{jiang2026mpdsgr}. Certified analysis establishes formal robustness guarantees for SNNs~\cite{mukhoty2024certified}. Temporal threat models expose failures in rate and timing information~\cite{hao2024threaten}, and combined attacks assess both channels jointly~\cite{wang2026robustsnnattacks}. This work addresses a different stage: the backbone is fixed, and only internal activity from the training split is used to calibrate the errors that remain after parameter-efficient adaptation.

\subsection{Parameter Efficient Adaptation and Firing Subspaces}

Bottleneck Adapters add residual down and up projections while retaining frozen backbone parameters~\cite{houlsby2019parameter}. LoRA instead learns low rank factors for selected weight updates~\cite{hu2022lora}. Both motivate lightweight adaptation, but neither specifies which temporal SNN responses should drive a post hoc correction. The proposed method uses an Adapter only to form the anchor predictor; its main representation is an error aligned subspace extracted from LIF firing curves.

The term \emph{firing scaffold} should not be confused with parameter pruning. Lottery ticket studies first identified sparse subnetworks in conventional neural networks~\cite{frankle2019lottery}, and related work extends this hypothesis to SNNs~\cite{kim2022lottery}. PAFS deletes no synaptic edge or channel: its scaffold comprises the dominant left singular directions of an aggregated firing--residual signature. It is a calibration representation, not a compressed replacement network.

\section{Method}
\label{sec:method}

\begin{figure*}[t]
\centering
\includegraphics[width=0.9\textwidth]{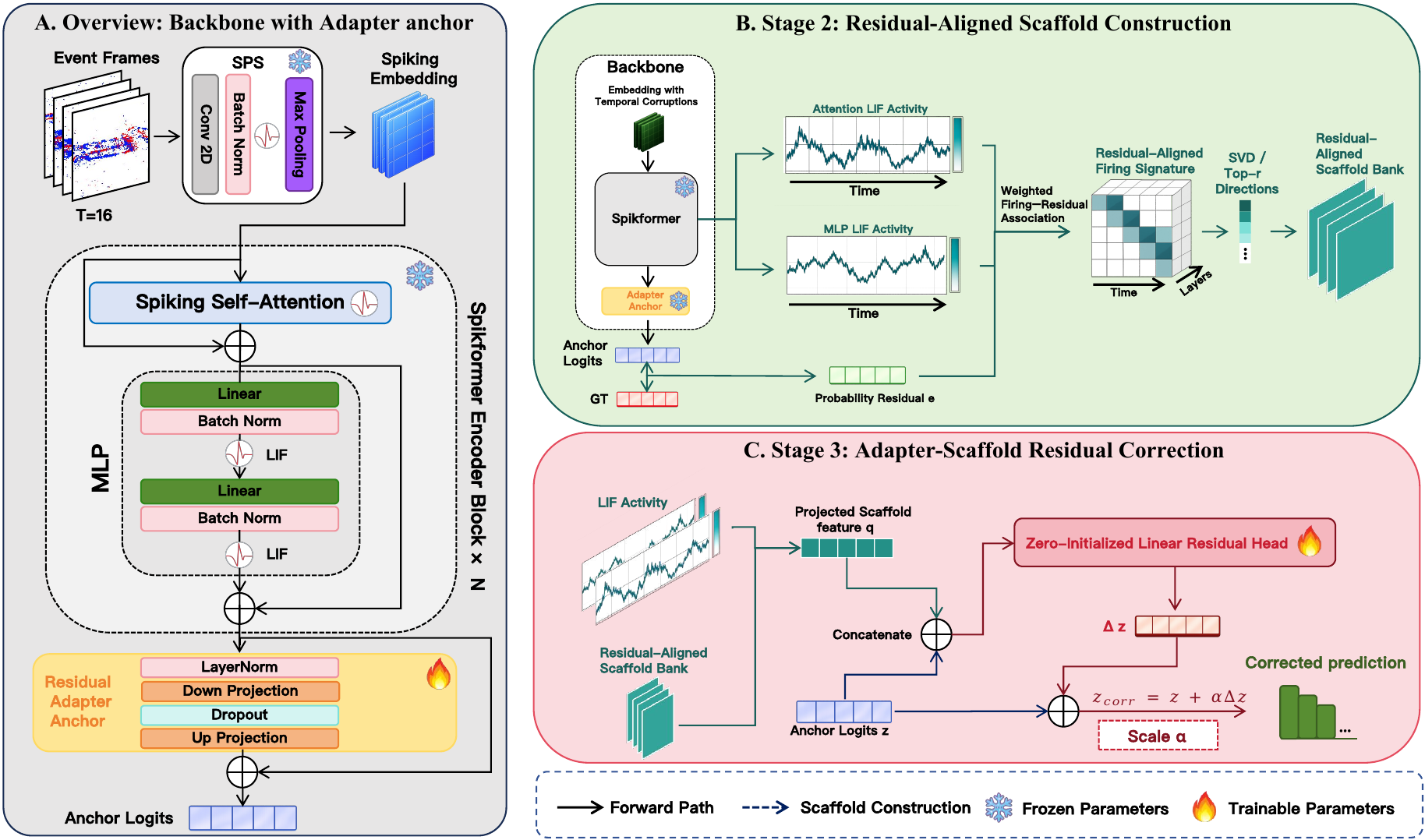}
\caption{
Overview of PAFS. The SNN backbone remains frozen throughout. Stage~1 learns a lightweight Adapter anchor under multiple temporal domains. Stage~2 constructs perturbation-aware firing scaffolds by aligning late-layer LIF trajectories with anchor prediction residuals and extracting dominant error-related firing directions. Stage~3 projects firing responses onto the fixed scaffolds and learns a residual logit correction for the anchor predictor.
}
\label{fig:framework}
\end{figure*}

\subsection{Overview}

Figure~\ref{fig:framework} summarizes the key idea of PAFS. Instead of modifying the pretrained backbone, PAFS treats the adapter predictor as an anchor and exploits residual-related temporal signals preserved in late-layer firing dynamics. The central design is to transform perturbation-induced firing responses into a compact residual-aligned scaffold, which provides complementary evidence for correcting the anchor prediction. After calibration, all learned components remain fixed, enabling label-free inference with a single backbone pass.

\subsection{Adapter Anchor}

Let $f_\theta$ denote a pretrained SNN with frozen backbone and classifier.
We define temporal domains as
\[
\mathcal P=\{p_0,p_1,\ldots,p_D\},
\]
where $p_0$ denotes the clean domain and the remaining domains correspond to temporal corruptions. For an input $x$, $\mathcal T_p(x)$ represents the transformation under domain $p$.
We first attach a lightweight Adapter after the final encoder block to obtain a fixed reference predictor. Let
$z_{\theta,\eta}(x)\in\mathbb R^K$
denote the logits produced by the SNN with frozen parameters $\theta$ and Adapter parameters $\eta$. Only Adapter parameters $\eta$ are optimized:
\begin{equation}
\mathcal L_{\rm adp}(\eta)
=
\mathbb E_{\substack{
(x,y)\sim\mathcal D_{\rm tr}\\
p\sim {\rm Unif}(\mathcal P)
}}
\left[
{\rm CE}
(
z_{\theta,\eta}(\mathcal T_p(x)),y
)
\right].
\label{eq:adapter_objective}
\end{equation}
We select the checkpoint $\eta^\star$ with the highest mean validation accuracy across all domains, freeze both $\theta$ and $\eta^\star$, and refer to $z_{\theta,\eta^\star}$ as the \emph{Adapter anchor}. The Adapter architecture, initialization, and parameter derivations
are provided in Appendix~A, while its optimization and model-selection
settings are summarized in Appendix~C.

\subsection{Residual-Aligned Firing Scaffold}

The central component of PAFS identifies late-layer firing directions associated with the remaining errors of the Adapter anchor. For selected LIF layers $\ell\in\mathcal L$, we use a training-only calibration subset \[ \mathcal D_{\rm cal} = \{(x_i,y_i)\}_{i=1}^{N_{\rm cal}} \subseteq \mathcal D_{\rm tr}. \] Validation and test samples are never used to estimate scaffold statistics or bases. For every sample and temporal domain, we collect the firing response
\begin{equation}
S_{\ell,i}^{p}
\in
\mathbb R^{T\times C_\ell\times N_\ell},
\end{equation}
where $\ell$ indexes the selected LIF layer, $i$ denotes the sample, and $p$ denotes the temporal domain. Here, $T$ is the number of time
steps, $C_\ell$ is the number of channels, and $N_\ell=H_\ell W_\ell$ is the number of spatial locations obtained by flattening the $H_\ell\times W_\ell$ feature map at layer $\ell$. Accordingly, $n\in\{1,\ldots,N_\ell\}$ indexes a flattened spatial location.

Since temporal corruptions primarily affect the evolution of evidence, we aggregate spatial tokens while preserving the temporal dimension:
\begin{equation}
R_{\ell,i}^{p}[c,t]
=
\frac{1}{N_\ell}
\sum_{n=1}^{N_\ell}
S_{\ell,i}^{p}[t,c,n].
\label{eq:firing_trajectory}
\end{equation}

The resulting representation $R_{\ell,i}^{p}\in\mathbb R^{C_\ell\times T}$ describes channel-wise temporal firing trajectories, where $c$ and $t$ index channel and time dimensions. 

To prevent intrinsically high-rate channels from dominating the subspace, we estimate $\mu_\ell$ and $\sigma_\ell$ over
$\mathcal D_{\rm cal}$ and all domains in $\mathcal P$, then standardize each trajectory:
\begin{equation}
\widetilde R_{\ell,i}^{p}
=
\frac{
R_{\ell,i}^{p}-\mu_\ell
}
{
\max(\sigma_\ell,10^{-4})
},
\label{eq:standardization}
\end{equation}
where $\mu_\ell,\sigma_\ell\in\mathbb R^{C_\ell\times T}$ are channel-time means and standard deviations. Subtraction, division, and the maximum in Eq.~\eqref{eq:standardization} are element-wise; the $10^{-4}$ floor prevents unstable scaling of nearly constant entries.

For the frozen Adapter anchor, define
\begin{equation}
\pi_i^p
=
{\rm softmax}\!\left(
z_{\theta,\eta^\star}(\mathcal T_p(x_i))
\right).
\label{eq:anchor_probability}
\end{equation}
Its probability residual is
\begin{equation}
e_i^p
=
\mathbf e_{y_i}-\pi_i^p,
\label{eq:residual}
\end{equation}
where $e_i^p\in\mathbb R^K$ is the residual vector, $\mathbf e_{y_i}\in\mathbb R^K$ is the one-hot label vector, and $K$ is the number of classes. Unlike a scalar prediction loss, $e_i^p$ preserves the class-wise direction of the remaining error: its ground-truth component indicates how much probability should be increased, while its negative components identify the over-predicted classes whose probabilities should be reduced. 

To emphasize the failure modes that remain unresolved by the Adapter anchor, we assign each calibration sample an error-aware weight:
\begin{equation}
w_i^p
=
1+\gamma\,
\mathbb{I}\!\left[
\arg\max_k \pi_i^p[k]\ne y_i
\right],
\qquad
\gamma\ge 0,
\label{eq:error_weight}
\end{equation}
where $\gamma$ controls the additional contribution of an incorrectly
classified sample. Thus, correctly classified samples retain unit weight,
whereas misclassified samples receive weight $1+\gamma$.

We next associate the internal firing response of each sample with the
class-wise correction required by the anchor. For a channel--time
coordinate $(c,t)$, $\widetilde R_{\ell,i}^p[c,t]$ measures the
standardized firing response, while $e_i^p[k]$ specifies the signed
probability adjustment required for class $k$. We therefore define the
sample-level firing--residual association
$A_{\ell,i}^p\in\mathbb R^{C_\ell\times(TK)}$ as
\begin{equation}
A_{\ell,i}^p[c,(t-1)K+k]
=
\widetilde R_{\ell,i}^p[c,t]\,
e_i^p[k],
\label{eq:sample_signature}
\end{equation}
where $t$ and $k$ index the temporal steps and classes, respectively. This construction pairs each channel--time firing response with each
class-wise residual component. A positive entry indicates that the corresponding channel--time response co-occurs with a requirement to increase the probability of class $k$, whereas a negative entry indicates a requirement to decrease it. Its magnitude reflects the joint strength of the firing response and the class-specific correction. Flattening the time--class pair is only a reshaping operation that preserves all temporal and class-wise associations while arranging firing channels as matrix rows.

For each temporal domain $p$, we aggregate these sample-level associations using their error-aware weights:
\begin{equation}
M_\ell^p
=
\frac{1}{
\sum_{i=1}^{N_{\rm cal}} w_i^p
}
\sum_{i=1}^{N_{\rm cal}}
w_i^p A_{\ell,i}^p
\in
\mathbb R^{C_\ell\times(TK)}.
\label{eq:signature}
\end{equation}
The rows of $M_\ell^p$ index firing channels, while its columns index time--class residual coordinates. Associations that recur across calibration samples with consistent signs accumulate in $M_\ell^p$, whereas sample-specific or inconsistent patterns tend to cancel. Consequently, $M_\ell^p$ acts as a weighted empirical cross-moment that summarizes which temporal channel responses are systematically associated with the anchor's remaining class-probability errors. It captures statistical association rather than causal attribution.

We define the set of corrupted domains as \[ \mathcal P_{\rm rob} = \mathcal P\setminus\{p_0\}. \] To prevent domains with larger firing-residual magnitudes from dominating the scaffold, we normalize each domain-specific signature before averaging: \begin{equation} \overline{M}_{\ell} = \frac{1}{|\mathcal P_{\rm rob}|} \sum_{p\in\mathcal P_{\rm rob}} \frac{M_{\ell}^{p}} {\max\!\left(\|M_{\ell}^{p}\|_{F},\epsilon_M\right)}, \label{eq:aggregation} \end{equation} where $\|\cdot\|_{F}$ denotes the Frobenius norm and $\epsilon_M=10^{-8}$ prevents division by zero for a degenerate domain-specific signature. This normalization gives each corruption domain comparable influence regardless of its overall association magnitude. The clean domain is excluded because the scaffold is designed to capture perturbation-induced firing--residual structure rather than nominal firing patterns.

Finally, we perform SVD:
\begin{equation}
\overline{M}_\ell
=
U_\ell \Sigma_\ell V_\ell^{\top}.
\label{eq:svd}
\end{equation}

Because $\overline{M}_\ell$ is organized in firing-channel space,
its left singular vectors define the dominant residual-aligned firing
directions. We retain the first $r$ directions:
\begin{equation}
B_\ell
=
U_{\ell,r}^{\top}
\in
\mathbb{R}^{r\times C_\ell},
\label{eq:basis}
\end{equation}
where $1 \le r \le \min(C_\ell,TK)$. The transpose converts the retained singular vectors into a projection matrix from channel space to scaffold coordinates. These rank-\(r\) directions maximize the Frobenius energy of the aggregated firing--residual association captured by the projection. The scaffold basis therefore preserves the dominant perturbation-related residual structure without modifying the backbone.

\subsection{Scaffold-based Residual Calibration}

Given an input $x$, the current firing trajectory is standardized using the stored calibration statistics $(\mu_\ell,\sigma_\ell)$ and projected onto the fixed scaffold:
\begin{subequations}
\label{eq:projection}
\begin{align}
Q_\ell(x)
&=
B_\ell \widetilde{R}_\ell(x)
\in
\mathbb{R}^{r\times T},
\label{eq:layer_projection}
\\
q(x)
&=
\operatorname{concat}_{\ell\in\mathcal{L}}
\operatorname{vec}\!\left(Q_\ell(x)\right)
\in
\mathbb{R}^{|\mathcal{L}|rT}.
\label{eq:scaffold_feature}
\end{align}
\end{subequations}
The projected feature \(q(x)\) measures the response of the current firing trajectory along the learned residual-aligned scaffold directions. It therefore summarizes whether the current sample exhibits internal temporal patterns that were associated with class-wise prediction residuals during scaffold construction.

The scaffold feature provides residual-related temporal information from the internal firing dynamics, while the Adapter anchor describes the current prediction through its logits, probabilities, and uncertainty statistics. We concatenate these complementary features to predict the residual logit correction:
\begin{equation}
\phi(z)
=
[z,\pi,\mathrm{conf}(z),
\mathrm{margin}(z),
\mathrm{ent}(z)]
\in
\mathbb R^{2K+3},
\label{eq:anchor_feature}
\end{equation}
where $z=z_{\theta,\eta^\star}(x), \pi={\rm softmax}(z).$

Writing \(\pi_{(1)}\) and \(\pi_{(2)}\) for the largest and second-largest
probabilities, the three scalar statistics are
\begin{subequations}
\label{eq:anchor_statistics}
\begin{align}
\mathrm{conf}(z)
&= \pi_{(1)},
\label{eq:anchor_conf}
\\
\mathrm{margin}(z)
&= \pi_{(1)}-\pi_{(2)},
\label{eq:anchor_margin}
\\
\mathrm{ent}(z)
&= -\frac{1}{\log K}
\sum_{k=1}^{K}\pi_k\log\pi_k .
\label{eq:anchor_entropy}
\end{align}
\end{subequations}
Confidence and margin summarize decision strength, while normalized entropy summarizes uncertainty; all three lie in \([0,1]\).

The correction head first predicts a logit residual:
\begin{equation}
\Delta z(x)
=
W\bigl[q(x);\phi(z(x))\bigr]
+
b_{\rm corr}
\in
\mathbb{R}^{K},
\label{eq:logit_residual}
\end{equation}
where $[\cdot;\cdot]$ denotes feature concatenation, $W\in\mathbb{R}^{K\times(|\mathcal{L}|rT+2K+3)}$, and $b_{\rm corr}\in\mathbb{R}^{K}$ are the trainable correction-head parameters. The corrected logits are then given by
\begin{equation}
z_{\rm corr}(x)
=
z(x)+\alpha\Delta z(x),
\label{eq:corrected_logits}
\end{equation}
where the scalar $\alpha$ controls the magnitude of the residual correction. We select $\alpha\in\{0.25,0.5,0.75,1.0\}$ according to the mean validation accuracy across the clean and corrupted domains, and keep the selected value fixed at test time.

Only the correction parameters $\omega=\{W,b_{\rm corr}\}$ are optimized, while the backbone parameters, Adapter parameters, and scaffold bases remain fixed. Writing \(x^p=\mathcal T_p(x)\), the
correction objective is
\begin{equation}
\begin{aligned}
\mathcal L_{\rm corr}
&=
\mathbb E_{\substack{
(x,y)\sim\mathcal D_{\rm tr}\\
p\sim{\rm Unif}(\mathcal P)
}}
\Big[
{\rm CE}(z_{\rm corr}(x^p),y)
\\[-1mm]
&\hspace{22mm}
+\lambda_\Delta\|\Delta z(x^p)\|_2^2
\Big].
\end{aligned}
\label{eq:correction_loss}
\end{equation}
Here, $\lambda_\Delta$ controls the penalty on unnecessary logit changes; we set \(\lambda_\Delta=10^{-3}\). Additional implementation details for firing hooks, tensor organization, and scaffold projection are provided in Appendix~A. The calibration protocol, correction-head optimization, and correction-scale selection are detailed in Appendix~C.

We initialize $W=b_{\rm corr}=0$, so correction training starts exactly from the Adapter anchor.

\subsection{Parameter and Inference Cost}

PAFS trains only the Adapter and correction head while keeping the pretrained predictor frozen. On DailyAction-DVS with Spikingformer, it adds 8.796K trainable parameters, corresponding to \(0.342\%\) of the 2.569M-parameter predictor; fixed scaffold buffers are excluded from this ratio. Detailed accounting is provided in Appendix~A, with measured efficiency results in Appendix~D. Inference requires one backbone pass, followed by scaffold projection and lightweight logit correction.
\section{Experiments}
\label{sec:experiments}
\begin{table*}[!t]
\centering
{\small
\renewcommand{\arraystretch}{0.92}
\setlength{\tabcolsep}{2.5pt}

\begin{tabular*}{\textwidth}
{@{\extracolsep{\fill}}lrrrrrr@{}}
\toprule
Method
& Clean
& Window Drop
& Jitter
& Speed
& Corr. Avg.
& Worst \\
\midrule


\multicolumn{7}{c}{\textbf{DailyAction-DVS, Spikformer}} \\
\midrule


LoRA
& \textbf{97.62$\pm$0.55}
& 86.51$\pm$1.40
& \textbf{91.35$\pm$2.10}
& 87.89$\pm$3.00
& 88.58$\pm$1.90
& 86.51$\pm$1.40 \\

Plain MLP 
& 97.52$\pm$0.60
& 86.16$\pm$1.55
& 89.62$\pm$2.35
& 87.89$\pm$3.20
& 87.89$\pm$2.05
& 86.16$\pm$1.55 \\

Adapter
& 97.35$\pm$0.53
& 86.62$\pm$1.22
& 90.43$\pm$2.50
& 86.85$\pm$3.65
& 87.97$\pm$1.57
& 86.62$\pm$1.22 \\

PAFS (ours)
& 97.46$\pm$0.53
& \textbf{87.20$\pm$1.51}
& 90.77$\pm$2.80
& \textbf{88.93$\pm$3.34}
& \textbf{88.97$\pm$2.28}
& \textbf{87.20$\pm$1.51} \\


\midrule
\multicolumn{7}{c}{\textbf{DailyAction-DVS, Spikingformer}} \\
\midrule


LoRA
& 98.62$\pm$0.40
& 91.35$\pm$2.80
& 93.43$\pm$1.90
& 92.04$\pm$2.75
& 92.27$\pm$1.85
& 91.35$\pm$2.80 \\

Plain MLP 
& 99.31$\pm$0.35
& 92.39$\pm$2.95
& 92.04$\pm$2.20
& 91.35$\pm$3.10
& 91.93$\pm$2.00
& 91.35$\pm$3.10 \\

Adapter
& 99.31$\pm$0.35
& 91.23$\pm$3.12
& 95.62$\pm$1.64
& 93.31$\pm$2.60
& 93.39$\pm$1.73
& 91.23$\pm$3.12 \\

PAFS (ours)
& \textbf{99.65$\pm$0.27}
& \textbf{94.69$\pm$2.03}
& \textbf{96.54$\pm$1.83}
& \textbf{94.46$\pm$2.96}
& \textbf{95.23$\pm$2.20}
& \textbf{94.46$\pm$2.96} \\


\midrule
\multicolumn{7}{c}{\textbf{CIFAR10-DVS, Spikingformer}} \\
\midrule


LoRA
& \textbf{58.80$\pm$1.10}
& 53.00$\pm$2.65
& 47.60$\pm$2.20
& 37.30$\pm$2.45
& 45.97$\pm$2.30
& 37.30$\pm$2.45 \\

Plain MLP 
& 56.00$\pm$1.25
& 52.30$\pm$2.90
& 45.80$\pm$2.40
& 29.40$\pm$3.10
& 42.50$\pm$2.55
& 29.40$\pm$3.10 \\

Adapter
& 57.50$\pm$1.32
& 53.43$\pm$2.84
& 48.80$\pm$1.95
& 37.60$\pm$2.43
& 46.61$\pm$2.34
& 37.60$\pm$2.43 \\

PAFS (ours)
& 57.43$\pm$1.21
& \textbf{53.53$\pm$3.00}
& \textbf{49.07$\pm$2.40}
& \textbf{37.73$\pm$2.66}
& \textbf{46.78$\pm$2.63}
& \textbf{37.73$\pm$2.66} \\


\midrule
\multicolumn{7}{c}{\textbf{DVS-GC, Spikformer}} \\
\midrule


LoRA
& \textbf{97.53$\pm$0.65}
& 33.80$\pm$3.10
& 56.07$\pm$7.50
& 79.34$\pm$4.20
& 58.40$\pm$4.55
& 33.80$\pm$3.10 \\

Plain MLP 
& 96.24$\pm$0.80
& 37.86$\pm$2.90
& 49.59$\pm$6.80
& 72.48$\pm$4.60
& 53.31$\pm$4.20
& 37.86$\pm$2.90 \\

Adapter
& 96.15$\pm$0.26
& 34.23$\pm$1.34
& 62.51$\pm$6.72
& 79.21$\pm$3.78
& 58.65$\pm$4.01
& 34.23$\pm$1.34 \\

PAFS (ours)
& 97.15$\pm$0.74
& \textbf{37.88$\pm$2.83}
& \textbf{66.74$\pm$8.62}
& \textbf{80.52$\pm$4.05}
& \textbf{61.71$\pm$4.45}
& \textbf{37.88$\pm$2.83} \\

\bottomrule
\end{tabular*}
}
\caption{
Robustness comparison across datasets and spiking Transformer backbones.
Results are mean test accuracy (\%) $\pm$ standard deviation over three
matched seeds; Corr. Avg. averages the three temporal corruptions, and
Worst denotes the lowest corrupted-condition accuracy.
}
\label{tab:main_results}
\end{table*}
\subsection{Protocol}

\textbf{Datasets and backbones.}
We evaluate PAFS on CIFAR10-DVS~\cite{li2017cifar10dvs}, DailyAction-DVS~\cite{liu2021event}, and the temporally composed DVS-Gesture-Chain (DVS-GC) benchmark~\cite{vicente2025spiking}.

To assess architectural transfer, we evaluate DailyAction-DVS with both Spikformer and Spikingformer~\cite{zhou2023spikformer,zhou2026spikingformer}. We use Spikingformer for CIFAR10-DVS and Spikformer for DVS-GC, yielding the four dataset--backbone settings reported in Table~\ref{tab:main_results}. The rationale for these dataset--backbone assignments, together with the complete input and temporal-corruption protocol, is provided in Appendix~B.

For the main evaluation, we use multiple dataset--backbone settings to examine whether PAFS generalizes across different spiking Transformer architectures. For controlled analyses, we select the backbone according to the objective of each study rather than enforcing a single backbone across all ablations. Mechanism comparisons are conducted with Spikingformer, which provides a strong backbone setting for evaluating whether the proposed residual-aligned representation contributes beyond the anchor prediction. Rank and channel-selection analyses are conducted with  Spikformer to study scaffold capacity and representation choices under a fixed architecture. Layer complementarity is further examined on both backbones to verify that the observed behavior is not architecture
specific.

All inputs are represented as $T=16$ event-count frames with a spatial resolution of $128\times128$. Within each setting, all methods use the same pretrained checkpoint, data splits, corruption settings, and evaluation
protocol.

\textbf{Corruption protocol.}
We evaluate clean inputs and three temporal corruptions: removing all events within a randomly selected interval covering 30\% of the sequence duration, perturbing event timestamps by offsets within \(\pm2\) temporal units, and rescaling timestamps to simulate a twofold speed change. All corrupted event streams are converted into \(T=16\) event-count frames using the same preprocessing as the clean inputs. Scaffold construction and correction-head training use only the training split. For PAFS and all trainable correction methods, the correction scale and all other hyperparameters are selected on the validation split, while their test split is used exclusively for final evaluation. The provenance of inherited Full checkpoints is reported separately in Appendix~B: two historical DVS-GC Full anchors and the historical Spikformer CIFAR10-DVS Full anchor were test-selected. They are used only as frozen starting points; no test examples are used to fit or select PAFS or the reported correction controls.

\textbf{Training and selection.}
The frozen backbone is adapted using clean inputs and the three corruptions with bottleneck width \(b=12\). PAFS uses two late LIF layers, error weight \(\gamma=4\), and rank \(r=4\) per layer. A training-only calibration subset is used for scaffold construction, with normalization statistics estimated without validation or test data. All newly trained Adapter, baseline, and correction checkpoints are selected according to mean validation accuracy across the four operating conditions, without using test data for fitting or selection. The inherited Full anchors follow the selection provenance reported in Appendix~B. Complete calibration splits, optimization schedules, hyperparameters,
software and hardware settings, and matched random seeds are provided
in Appendix~C.

\textbf{Compared methods.} For the main robustness comparison in Table~\ref{tab:main_results}, we compare PAFS with LoRA~\cite{hu2022lora}, the Adapter anchor, and Plain MLP on Full. Plain MLP denotes a two-layer residual MLP attached to the logit descriptor of the original frozen checkpoint, without using an Adapter anchor or internal firing features. Its parameter budget is approximately matched to that of the Adapter.

For the controlled representation analysis in Table~\ref{tab:basis_controls}, we additionally evaluate a Logit-only MLP. This baseline operates on the frozen Adapter anchor, receives only the anchor logit descriptor $\phi(z)$, and follows the same second-stage training objective and validation-based correction-scale selection as PAFS. Its trainable parameter budget is matched to that of the PAFS correction head. Architectural and optimization details for all baselines and controls are provided in Appendix~C.

\textbf{Metrics.}
We report clean accuracy, corrupted-average accuracy over Window Drop, Jitter, and Speed, and Worst, defined as the lowest cross-seed mean accuracy among the three corruptions. Results are reported as mean \(\pm\) standard deviation over three matched runs. 
\begin{table*}[!t]
\centering
{\small
\renewcommand{\arraystretch}{0.90}
\setlength{\tabcolsep}{2.2pt}

\begin{tabular*}{\textwidth}
{@{\extracolsep{\fill}}llrrrrrrr@{}}
\toprule
Dataset
& Representation
& Clean
& Window Drop
& Jitter
& Speed
& Avg.
& Corr. Avg.
& Worst
\\
\midrule

\multirow{5}{*}{CIFAR10-DVS}
& PAFS (ours)
& 57.43$\pm$1.21
& \textbf{53.53$\pm$3.00}
& \textbf{49.07$\pm$2.40}
& \textbf{37.73$\pm$2.66}
& \textbf{49.44$\pm$2.63}
& \textbf{46.78$\pm$2.63}
& \textbf{37.73$\pm$2.66}
\\

& PCA scaffold
& 56.93$\pm$1.08
& 52.87$\pm$2.55
& 47.42$\pm$2.08
& 36.31$\pm$2.18
& 48.38$\pm$1.92
& 45.53$\pm$2.03
& 36.31$\pm$2.18
\\

& Random scaffold
& 56.84$\pm$1.17
& 52.61$\pm$2.72
& 47.18$\pm$2.22
& 36.74$\pm$2.36
& 48.34$\pm$2.06
& 45.51$\pm$2.14
& 36.74$\pm$2.36
\\

& Logit-only MLP
& 56.71$\pm$1.12
& 52.76$\pm$2.63
& 47.05$\pm$1.96
& 36.58$\pm$2.27
& 48.28$\pm$1.98
& 45.46$\pm$2.07
& 36.58$\pm$2.27
\\

& Adapter anchor
& \textbf{57.50$\pm$1.32}
& 53.43$\pm$2.84
& 48.80$\pm$1.95
& 37.60$\pm$2.43
& 49.33$\pm$2.34
& 46.61$\pm$2.34
& 37.60$\pm$2.43
\\

\midrule

\multirow{5}{*}{DailyAction-DVS}
& PAFS (ours)
& \textbf{99.65$\pm$0.27}
& \textbf{94.69$\pm$2.03}
& \textbf{96.54$\pm$1.83}
& \textbf{94.46$\pm$2.96}
& \textbf{96.34$\pm$2.20}
& \textbf{95.23$\pm$2.20}
& \textbf{94.46$\pm$2.96}
\\

& PCA scaffold
& 99.54$\pm$0.24
& 93.15$\pm$2.36
& 95.08$\pm$1.86
& 93.38$\pm$2.64
& 95.29$\pm$1.88
& 93.87$\pm$2.07
& 93.15$\pm$2.36
\\

& Random scaffold
& 99.42$\pm$0.31
& 92.47$\pm$2.58
& 95.36$\pm$1.74
& 92.28$\pm$2.76
& 94.88$\pm$1.96
& 93.37$\pm$2.12
& 92.28$\pm$2.76
\\

& Logit-only MLP
& 99.31$\pm$0.35
& 91.82$\pm$2.68
& 94.39$\pm$2.08
& 91.47$\pm$2.94
& 94.25$\pm$2.08
& 92.56$\pm$2.24
& 91.47$\pm$2.94
\\

& Adapter anchor
& 99.31$\pm$0.35
& 91.23$\pm$3.12
& 95.62$\pm$1.64
& 93.31$\pm$2.60
& 94.87$\pm$1.73
& 93.39$\pm$1.73
& 91.23$\pm$3.12
\\

\bottomrule
\end{tabular*}
}
\caption{
Mechanism controls with Spikingformer under a matched Adapter anchor and correction protocol. All variants differ only in the representation used for correction. Results are mean test accuracy (\%) $\pm$ standard deviation over three matched seeds.
}
\label{tab:basis_controls}
\end{table*}

\subsection{Robustness Across Backbones}
Table~\ref{tab:main_results} shows that PAFS achieves the highest reported mean Corr. Avg. and Worst accuracies across all four dataset--backbone settings. Its advantage over the corresponding Adapter anchor is observed across all evaluated corruptions rather than being concentrated in a particular perturbation or architecture, indicating that the proposed correction is broadly applicable within the studied deployment setting. This consistency suggests that the PAFS design remains effective across different dataset characteristics and backbone-specific temporal computations without relying on a single failure pattern.

The magnitude of improvement is nevertheless task dependent. PAFS provides clearer gains on DVS-GC and DailyAction-DVS, where recognition relies more strongly on temporal structure, but offers only a marginal improvement on CIFAR10-DVS. This pattern suggests that the method is most useful when the errors remaining after Adapter training are still reflected in late-layer firing trajectories. Moreover, PAFS consistently outperforms the Logit-only MLP, although both use the same Adapter anchor and residual-correction objective. The comparison supports the conclusion that internal firing dynamics contain correction-relevant information that is not fully represented by the anchor logits alone.

\begin{figure}[ht]
\centering

\includegraphics[
    width=\columnwidth,
    keepaspectratio
]{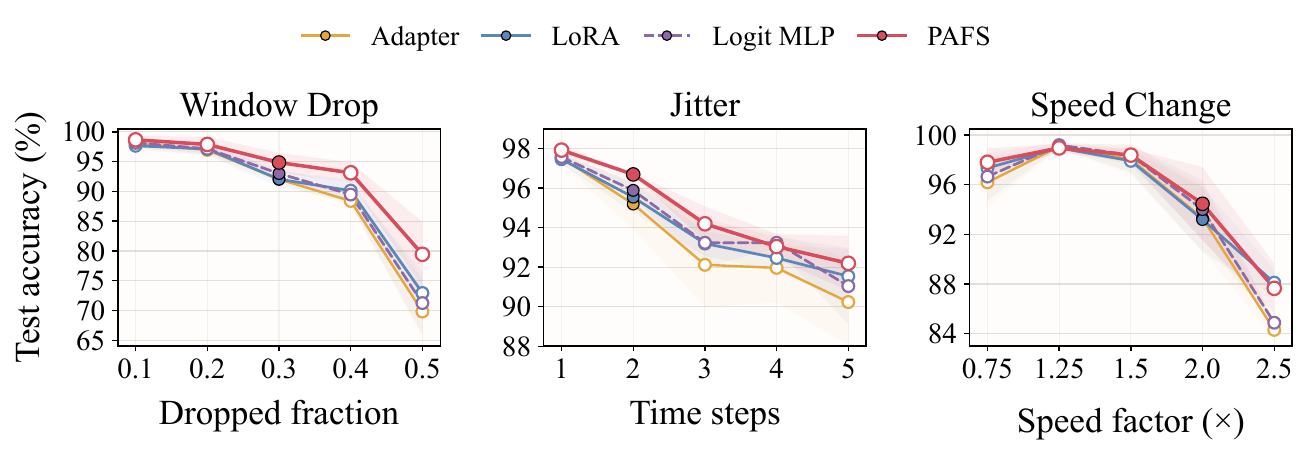}
\caption{
Generalization across Window Drop, Jitter, and Speed severities on DailyAction-DVS with Spikingformer. Results are averaged over three matched random seeds, with shaded regions showing $\pm1$ standard deviation.
}
\label{fig:severity_generalization}
\end{figure}

\textbf{Generalization to unseen corruption severities.} Figure~\ref{fig:severity_generalization} evaluates all methods across corruption strengths beyond the nominal training settings. PAFS retains
the highest or competitive accuracy over most Window Drop and Jitter levels, with its clearest advantage under severe event removal. Under Speed Change, performance degrades toward both extreme speed factors; PAFS remains competitive around the nominal operating range but does not dominate every unseen severity. These results indicate that the residual-aligned scaffold transfers across perturbation magnitudes, although the benefit is non-uniform across corruption types.

\subsection{Representation and Design Analysis}\textbf{Representation controls.} Table~\ref{tab:basis_controls} compares PAFS with matched logit-only, PCA-scaffold, and random-scaffold variants under the same Adapter anchor and correction protocol. PAFS achieves the strongest robustness, with a clearer advantage on DailyAction-DVS, particularly under Window Drop, while differences on CIFAR10-DVS are modest. The improvement over the logit-only variant shows that late-layer firing trajectories contain correction-relevant information beyond the anchor logits. Its advantage over PCA and random bases further shows that access to internal activations or low-rank compression alone is insufficient; aligning firing directions with the Adapter's class-wise probability residuals yields a representation better matched to the Adapter anchor's remaining temporal errors. Appendix~D further examines sensitivity to the number of training-only calibration samples and compares PAFS with scalar channel-ranking criteria.

\begin{figure}[ht]
\centering
\includegraphics[width=\linewidth]
{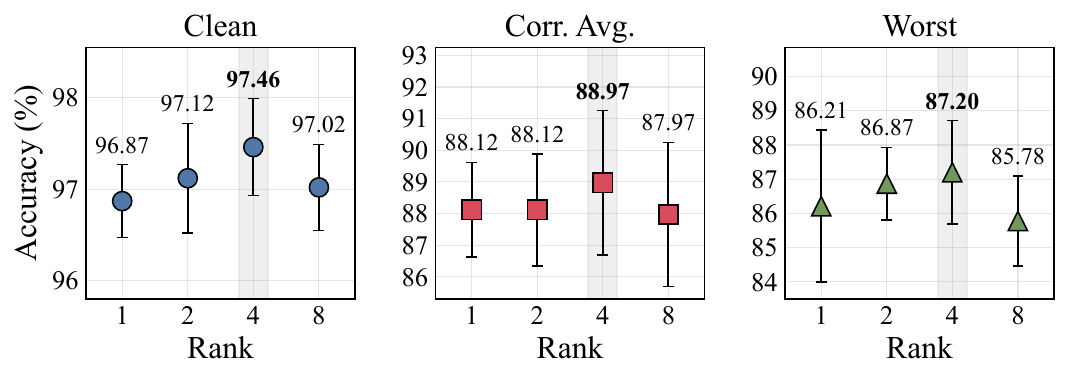}
\caption{
Rank sensitivity of PAFS on DailyAction-DVS with Spikformer. The dashed line marks the validation-selected rank \(r=4\); error bars show \(\pm1\) standard deviation over three matched seeds.
}

\label{fig:rank}
\end{figure}

\textbf{Rank sensitivity.} Figure~\ref{fig:rank} shows that PAFS remains broadly stable across different scaffold ranks, with \(r=4\) selected by validation performance. Increasing the rank does not consistently improve robustness, suggesting that the dominant residual-related firing structure is compact and that additional directions mainly introduce redundant variation. These results support the use of a small low-rank scaffold for temporal robustness calibration. 

\begin{figure}[!t]
\centering
\includegraphics[
    width=\columnwidth,
    keepaspectratio
]{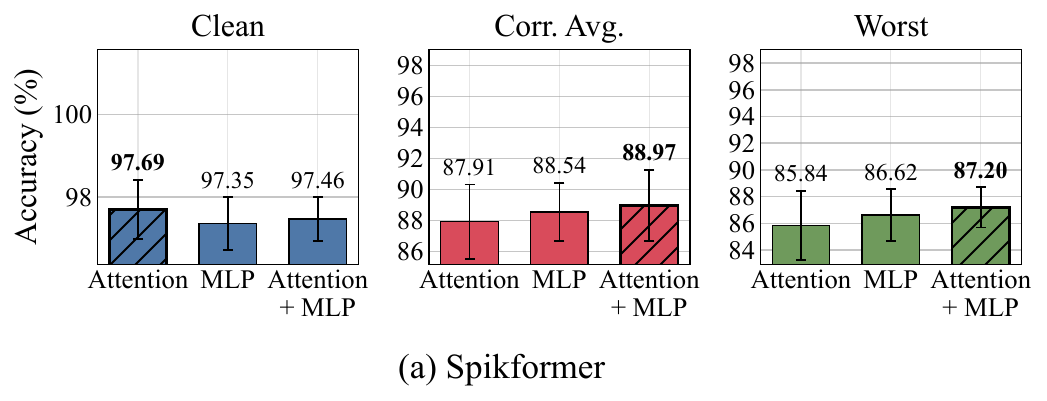}
\par
\includegraphics[
    width=\columnwidth,
    keepaspectratio
]{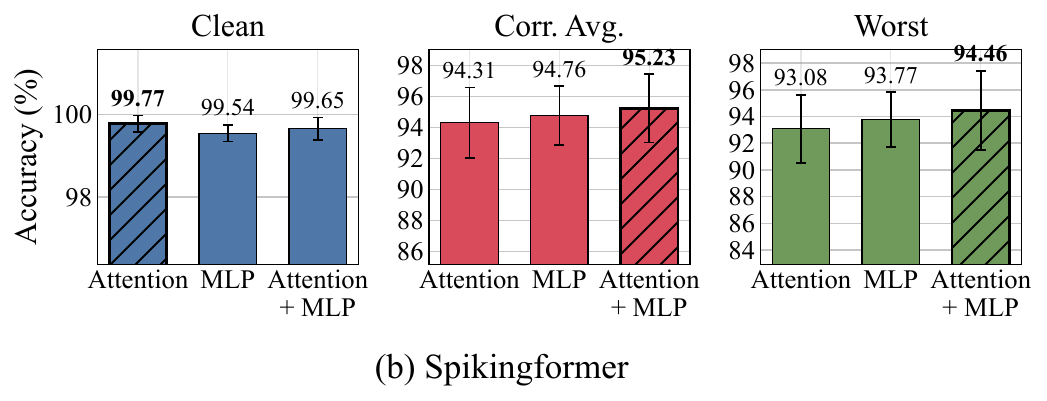}

\caption{
Layer evidence on DailyAction-DVS with Spikformer and Spikingformer. Attention+MLP denotes the final PAFS configuration; error bars show $\pm1$ standard deviation over three matched seeds.
}
\label{fig:layer}
\end{figure}
\textbf{Layer complementarity.}
Figure~\ref{fig:layer} compares attention-only, MLP-only, and combined attention--MLP scaffolds under the same validation-based protocol. The combined scaffold achieves the strongest robustness across both backbones, while attention-only features better preserve clean accuracy. This consistent behavior suggests that attention and MLP firing responses capture different but complementary aspects of the residual error structure, and their combination provides more effective evidence for residual correction.

\section{Conclusion}
PAFS extracts residual-aligned firing subspaces from a frozen SNN for post-training calibration. It improves robustness over Adapter and Logit-only correction without modifying the predictor or adding a backbone pass.

\FloatBarrier
\clearpage
\putbib[paper/references]
\end{bibunit}
\clearpage
\begin{bibunit}[pafs-author-year]
\pafsbibcontext{.supp}
\renewcommand{\theequation}{S\arabic{equation}}
\renewcommand{\thefigure}{S\arabic{figure}}
\renewcommand{\thetable}{S\arabic{table}}
\setcounter{equation}{0}
\renewcommand{\theHequation}{S.\arabic{equation}}
\renewcommand{\theHfigure}{S.\arabic{figure}}
\renewcommand{\theHtable}{S.\arabic{table}}
\providecommand{\theHalgorithm}{\arabic{algorithm}}
\renewcommand{\theHalgorithm}{S.\arabic{algorithm}}
\setcounter{figure}{0}
\setcounter{table}{0}

\begin{center}
\large \textbf{Supplementary Material for}\\
\normalsize \textit{PAFS: Perturbation-Aware Firing Scaffolds for Robust Event-Based SNNs}
\end{center}

This supplement provides additional derivations, dataset rationale,
reproducibility settings, and experimental analyses supporting the main paper.
We focus on architectural details, parameter and memory accounting,
experimental protocols, and implementation settings required to reproduce
and interpret the results.

\section{Appendix A: Additional Derivations and Method Details}
\label{sec:supp_derivations}

\paragraph{Adapter architecture.}
The Adapter is inserted after the final encoder block of the frozen SNN. It uses
a residual bottleneck parameterization~\citep{houlsby2019parameter}. For a
channel-last feature tensor $h\in\mathbb R^{T\times N\times C}$, the same
bottleneck transformation is shared over all temporal steps and spatial
positions:
\begin{equation}
\begin{aligned}
u_{t,n}
&=
W_{\rm down}
{\rm LN}_{\gamma,\beta}(h_{t,n})
+b_{\rm down},
\\
\mathcal A_{\eta}(h)_{t,n}
&=
h_{t,n}
+
W_{\rm up}\,
{\rm Drop}\!\left(
{\rm GELU}(u_{t,n})
\right)
+b_{\rm up}.
\end{aligned}
\label{eq:supp_adapter}
\end{equation}
where $W_{\rm down}\in\mathbb R^{b\times C}$ and
$W_{\rm up}\in\mathbb R^{C\times b}$. Here, $\mathrm{LN}$ denotes Layer
Normalization~\citep{ba2016layernorm}, and GELU denotes the Gaussian Error
Linear Unit~\citep{hendrycks2016gelu}. The trainable parameter set contains the
down- and up-projection weights, their biases, and the LayerNorm affine
parameters. The up-projection weight and bias are initialized to zero, so the
Adapter branch initially contributes no residual and the predictor starts
exactly from the original frozen SNN.

\paragraph{Complete calibration procedure.}
Algorithm~S1 summarizes the complete PAFS training and deployment
procedure. From the checkpoint training indices, we form a fixed training-only
calibration pool
\[
\mathcal D_{\mathrm{pool}}
=
\mathcal D_{\mathrm{fit}}
\mathbin{\dot\cup}
\mathcal D_{\mathrm{rest}}
\subset
\mathcal D_{\mathrm{tr}},
\]
where
\(
|\mathcal D_{\mathrm{fit}}|=384
\)
and
\(
|\mathcal D_{\mathrm{rest}}|=128
\)
under the default configuration.

Only \(\mathcal D_{\mathrm{fit}}\) is used to estimate the
channel--time normalization statistics and construct the
residual-aligned scaffold bases. The remaining
\(\mathcal D_{\mathrm{rest}}\) subset is excluded from these
Stage~2 operations. This partition does not define a
train/validation split: both subsets remain part of
\(\mathcal D_{\mathrm{tr}}\) and may appear as ordinary training
examples during Adapter and correction-head optimization.
The \(\mathcal D_{\mathrm{rest}}\) subset is not used as validation
data for checkpoint or deployment-scale selection, or for final
test reporting.

\renewcommand{\thealgorithm}{S\arabic{algorithm}}
\setcounter{algorithm}{0}
\begin{algorithm}[!bt]
\caption{PAFS calibration and prediction}
\label{alg:pafs}
\begin{algorithmic}[1]

\REQUIRE Frozen SNN $f_\theta$; training set $\mathcal D_{\rm tr}$;
scaffold-fitting subset $\mathcal D_{\rm fit}$;
temporal domains $\mathcal P$; selected LIF layers $\mathcal L$;
scaffold rank $r$

\ENSURE Adapter anchor $\eta^\star$;
fixed scaffold buffers
$\{B_\ell,\mu_\ell,\sigma_\ell\}_{\ell\in\mathcal L}$;
trained correction head $\omega^\star$;
deployment scale $\hat{\alpha}$

\STATE \textbf{Stage 1: Train the Adapter anchor}

\STATE Initialize Adapter parameters $\eta$

\FOR{each Adapter update}

\STATE Sample $(x,y)\sim\mathcal D_{\rm tr}$ and
$p\sim{\rm Unif}(\mathcal P)$

\STATE Set $x^p\leftarrow\mathcal T_p(x)$

\STATE Update $\eta$ using
$\mathrm{CE}(z_{\theta,\eta}(x^p),y)$

\ENDFOR

\STATE Select validation-best checkpoint $\eta^\star$ and freeze it

\STATE \textbf{Stage 2: Construct residual-aligned scaffolds}

\FOR{$(x_i,y_i)\in\mathcal D_{\rm fit}$}
    \FOR{$p\in\mathcal P$}
        \STATE Run the frozen anchor on
        $x_i^p=\mathcal T_p(x_i)$

        \STATE Collect trajectories
        $\{R_{\ell,i}^{p}\}_{\ell\in\mathcal L}$
        and anchor probability $\pi_i^p$

        \STATE Compute residual $e_i^p$
        and error-aware weight $w_i^p$
    \ENDFOR
\ENDFOR

\FOR{$\ell\in\mathcal L$}

\STATE Estimate $\mu_\ell,\sigma_\ell$ and standardize trajectories

\STATE Construct weighted firing-residual matrices

\STATE Normalize and aggregate the corrupted-domain matrices

\STATE Compute SVD

\STATE Store scaffold basis $B_\ell$

\ENDFOR

\STATE Freeze scaffold bases and normalization statistics

\STATE \textbf{Stage 3: Train the correction head at unit scale}
\STATE Initialize correction head $\omega$ to zero

\FOR{each correction-head update}
    \STATE Sample $(x,y)\sim\mathcal D_{\rm tr}$ and
    $p\sim{\rm Unif}(\mathcal P)$
    \STATE Set $x^p\leftarrow\mathcal T_p(x)$
    \STATE Run the frozen Adapter anchor once on $x^p$
    \STATE Project the selected firing trajectories onto
    $\{B_\ell\}_{\ell\in\mathcal L}$
    \STATE Compute the raw residual $\Delta z_\omega(x^p)$
    \STATE Set
    $z^{\rm train}(x^p)\leftarrow
    z(x^p)+\Delta z_\omega(x^p)$
    \STATE Update only $\omega$ using Eq.~(18) of the main paper
    with $\alpha_{\rm train}=1$
\ENDFOR

\STATE Freeze the trained correction head as $\omega^\star$

\STATE \textbf{Validation-only scale selection}
\STATE Set
$\mathcal A_{+}=\{0.25,0.5,0.75,1.0\}$
and
$\mathcal A=\{0\}\cup\mathcal A_{+}$
\STATE Evaluate $\mathcal A$ on the validation split
\STATE Retain scales satisfying the clean-performance guard
\STATE Select the retained scale $\hat{\alpha}$ with the highest
validation robust average

\STATE \textbf{Inference:}
run the frozen anchor once and output
$z^{\rm corr}=z+\hat{\alpha}\Delta z_{\omega^\star}$

\end{algorithmic}
\end{algorithm}

\paragraph{Scaffold and correction implementation.}
We follow Eqs.~(2)--(18) of the main paper for firing extraction, spatial
aggregation, channel--time standardization, probability residuals,
error-aware weighting, domain-balanced association matrices, SVD-based scaffold
construction, scaffold projection, and residual correction. For each selected
source, the hooked LIF output is arranged as a
$T\times C_{\ell}\times N_{\ell}$ tensor, where the spatial dimensions are
flattened into $N_{\ell}$. Spatial locations are averaged while time is
preserved, producing a $C_{\ell}\times T$ trajectory for each sample and domain.
The two selected late-layer sources are the attention projection LIF response
(\textrm{proj\_lif}) and the first MLP LIF response
(\textrm{fc1\_lif}). The bases and normalization statistics are frozen after
construction, and the correction head operates on their projected coordinates
without an additional backbone pass.

\paragraph{Relation to probability calibration.}
PAFS learns a feature-dependent residual logit correction for robustness; it
does not claim calibrated confidence probabilities. This distinction matters
because conventional post-hoc probability calibration aims to align predicted
confidence with empirical correctness~\citep{guo2017calibration}, whereas PAFS
uses firing trajectories to correct perturbation-induced classification errors.

\paragraph{Trainable-parameter derivation.}
For hidden dimension $C$ and bottleneck dimension $b$, the Adapter contains
\begin{equation}
N_{\rm adapter}=2Cb+b+3C
\label{eq:supp_adapter_parameters}
\end{equation}
trainable parameters. The two projection matrices contribute $2Cb$, their
biases contribute $b+C$, and the LayerNorm affine parameters contribute $2C$.
The correction feature has dimension $|\mathcal L|rT+2K+3$. Including the
output bias, the correction head contains
\begin{equation}
N_{\rm corr}
=
K\left(|\mathcal L|rT+2K+4\right),
\label{eq:supp_correction_parameters}
\end{equation}
and the total learned overhead is
\begin{equation}
N_{\rm add}=N_{\rm adapter}+N_{\rm corr}.
\label{eq:supp_total_parameters}
\end{equation}
For DailyAction-DVS with Spikingformer,
$C=256$, $b=12$, $K=12$, $|\mathcal L|=2$, $r=4$, and $T=16$, giving
\begin{equation*}
\begin{aligned}
N_{\rm adapter} &= 6{,}924,\\
N_{\rm corr}    &= 1{,}872,\\
N_{\rm add}     &= 8{,}796.
\end{aligned}
\end{equation*}
This is approximately $0.342\%$ of the frozen 2.569M-parameter predictor.

\paragraph{Persistent-buffer derivation.}
The fixed scaffold bases and channel--time normalization statistics require
\begin{equation}
N_{\rm basis}=r\sum_{\ell\in\mathcal L}C_{\ell},
\qquad
N_{\rm norm}=2T\sum_{\ell\in\mathcal L}C_{\ell}
\label{eq:supp_buffer_parameters}
\end{equation}
stored scalars. For two selected layers with $C_{\ell}=256$, $r=4$, and
$T=16$, the bases contain 2,048 scalars and the normalization statistics contain
16,384 scalars. In FP32, these fixed buffers require 72.0~KiB. They receive no
gradients and are excluded from the trainable-parameter percentage.

\FloatBarrier

\section{Appendix B: Dataset Selection and Evaluation Protocol}
\label{sec:supp_datasets}

\paragraph{Evaluation Scope.}
The evaluation is organized around three claims. First, PAFS should remain
effective across event-recognition tasks with different dependence on temporal
structure. Second, its gains should transfer across spiking-transformer
architectures rather than depend on one particular backbone implementation.
Third, the method should improve robustness under complementary temporal failure
modes without sacrificing performance on clean inputs. The dataset--backbone
settings, corruption suite, and evaluation metrics are selected to test these
claims separately.

\subsection{Datasets}

\paragraph{CIFAR10-DVS.}
CIFAR10-DVS is a balanced event-based object-classification benchmark
containing 10,000 recordings from the ten CIFAR-10 categories, with
1,000 event streams per class~\citep{li2017cifar10dvs}. The event streams are
generated by displaying static CIFAR-10 images on a monitor and moving them
along predefined trajectories while a $128\times128$ DVS records the resulting
intensity changes. Consequently, the semantic label is inherited from static
object identity rather than from naturally evolving motion.

We use CIFAR10-DVS as a weak temporal-order control. It tests whether PAFS
remains useful when long-range temporal composition is not itself
label-defining, while its display-generated motion means that it is not used as
the primary evidence for robustness to natural human actions. The release does
not prescribe one unique train/validation/test partition, so the split used in
this work is treated as part of the experimental protocol and is kept identical
across all compared methods.

\paragraph{DailyAction-DVS.}
DailyAction-DVS contains 1,440 directly recorded event clips of 15 subjects
performing 12 daily actions: bend, climb, fall down, get up, jump, lie down,
carry box, run, sit down, stand up, walk, and pick up
~\citep{liu2021event}. Recordings cover two illumination conditions and two
camera placements with different viewing distances and angles. Each example is
a short, cropped, single-action clip rather than a continuous activity stream.

Unlike display-converted object datasets, DailyAction-DVS directly records
human motion and includes variation in subject, illumination, and viewpoint.
We therefore use it as the principal setting for evaluating robustness to
temporal perturbations in natural action recordings. DailyAction-DVS is also
evaluated with both Spikformer and Spikingformer, allowing architectural
transfer to be studied while holding the task, split, preprocessing, and
corruption protocol fixed.

\paragraph{DVS-Gesture-Chain.}
DVS-Gesture-Chain (DVS-GC) is an order-sensitive benchmark generated from the
real event-camera recordings of DVS128 Gesture rather than an independently
captured corpus~\citep{amir2017low,vicente2025spiking}. The upstream DVS128 Gesture
dataset contains 1,342 labeled gesture instances from 29 subjects under three
illumination conditions. DVS-GC constructs examples by selecting gesture
primitives and concatenating them into ordered chains.

Different orders are assigned different labels, so two samples containing the
same local gesture primitives may belong to different classes solely because
their temporal arrangements differ. DVS-GC therefore provides the most explicit
test of whether PAFS preserves structured timing information rather than only
aggregate firing activity or event counts. Because the chains are
programmatically composed, we interpret DVS-GC as a controlled temporal-order
benchmark rather than as a dataset of naturally occurring long activities. The audited configuration uses $T=16$, 81 classes, train/validation/test sizes of 6,318/1,620/1,944, and a shard size of 64. The generator source exposes a composition parameter $\alpha_{\rm gen}\in\{0.5,0.7\}$ and seed~123 as defaults, but the cached artifacts do not retain these values. We therefore do not present them as verified run-level settings.
\paragraph{Why the datasets are complementary.}
The three datasets are selected to test PAFS under progressively stronger
forms of temporal dependence. CIFAR10-DVS provides an object-recognition
setting in which class identity is largely determined by accumulated spatial
evidence and temporal order is comparatively weak. DailyAction-DVS introduces
natural human-action dynamics, where recognition depends more strongly on how
motion evidence evolves over time. DVS-GC further provides an explicitly
order-sensitive setting in which similar elementary gestures form different
classes through temporal composition. This progression is directly aligned
with the central hypothesis of PAFS: residual errors should be more recoverable
from late-layer firing trajectories when task-relevant temporal structure
remains encoded beyond the anchor logits.

\paragraph{Dataset papers and official resources.}
The dataset papers and official or author-provided resources used in this work
are listed below.

\begin{itemize}
\item \textbf{CIFAR10-DVS:}
dataset paper:
\url{https://doi.org/10.3389/fnins.2017.00309};
official data release:
\url{https://doi.org/10.6084/m9.figshare.4724671.v2}.

\item \textbf{DailyAction-DVS:}
dataset paper:
\url{https://www.ijcai.org/proceedings/2021/240};
author repository and data instructions:
\url{https://github.com/qianhuiliu/SNN-action-recognition}.

\item \textbf{DVS-Gesture-Chain:}
dataset paper:
\url{https://doi.org/10.1016/j.neucom.2024.128657};
official generator:
\url{https://github.com/VicenteAlex/DVS-Gesture-Chain}.
The benchmark is constructed from DVS128 Gesture, whose original paper is
cited in the main text.

\end{itemize}

\subsection{Backbone Architectures}

\paragraph{Shared architecture and input convention.}
Every audited backbone receives a mini-batch of event-frame sequences with shape
$[B,T,2,128,128]$, where $T=16$ and the two input channels encode event
polarity. Both encoders have depth~2, hidden dimension~256, MLP hidden
dimension~1,024, and patch size~16. The patch embedding therefore yields an
$8\times8$ token grid with 64 spatial tokens per simulation step. The temporal
dimension is retained throughout the multistep spiking layers: spatial token
interaction is computed within each step, whereas membrane states propagate
information across time.

\paragraph{Spikformer.}
Spikformer is a directly trained spiking vision Transformer that replaces
conventional floating-point self-attention with Spiking Self-Attention
(SSA)~\citep{zhou2023spikformer}. The audited implementation uses fixed LIF
neurons with $\tau=2.0$ and the PyTorch backend. Its architecture consists of a
Spiking Patch Splitting stem, a sequence of Transformer encoder blocks
containing SSA and spiking MLP modules, global average pooling, and a linear
classifier. Query,
key, and value representations are converted into spike-form tensors, and the
attention interaction is performed without softmax.

Spikformer uses a post-activation, spike-element-wise residual organization:
the output of a spiking branch is added to the shortcut representation. It
therefore supplies PAFS with firing trajectories generated by a canonical
softmax-free spiking Transformer with attention and MLP pathways.

\paragraph{Spikingformer.}
Spikingformer retains softmax-free global spike attention but changes the
placement of spiking activations and the organization of residual propagation
~\citep{zhou2026spikingformer}. The audited model is the base
ConvBN$\rightarrow$LIF$\rightarrow$MaxPool tokenizer variant, rather than a
dagger or CML-downsampling variant. It uses fixed LIF neurons with $\tau=1.5$
and the CuPy backend, followed by pre-activation Spiking Self-Attention and
pre-activation spiking MLP blocks.
Before a learned projection, the residual representation is converted into
binary spikes, while the membrane-valued shortcut is propagated across blocks.

The central distinction can be summarized as
\begin{equation}
\underbrace{\operatorname{SN}(\mathcal F(X))+X}_{\text{Spikformer}}
\qquad\text{versus}\qquad
\underbrace{\mathcal F(\operatorname{SN}(X))+X}_{\text{Spikingformer}},
\label{eq:supp_residual_comparison}
\end{equation}
where $\mathcal F$ denotes a learned transformation and
$\operatorname{SN}$ denotes a spiking-neuron operation. Spikingformer therefore
changes the activation placement, residual information path, and internal
firing organization rather than merely changing the parameter count.

\paragraph{Why both backbones are evaluated.}
Both models use spiking-neuron dynamics, spike-form attention representations,
softmax-free global token interaction, and attention--MLP encoder blocks.
However, their tokenizers, projection operators, activation placement, and
residual representations differ. Spikformer evaluates PAFS under a
post-activation spike-element-wise shortcut, whereas Spikingformer evaluates it
under a pre-activation membrane-shortcut design in which learned
transformations receive binary spike inputs.

DailyAction-DVS is evaluated with both backbones while the dataset split, event
conversion, temporal corruptions, and model-selection protocol remain fixed.
CIFAR10-DVS is additionally evaluated with Spikingformer, and DVS-GC with
Spikformer, producing four dataset--backbone settings. Consistent behavior
across these settings supports cross-backbone transfer within spiking
Transformers; it does not imply generalization to every possible SNN
architecture.

\paragraph{How PAFS interfaces with the backbones.}
PAFS leaves the pretrained backbone and original classifier frozen. The
Adapter is inserted after the final encoder block, while firing trajectories
are collected from two late internal sources: the attention projection LIF
response (\textrm{proj\_lif}) and the first MLP LIF response
(\textrm{fc1\_lif}). Each captured response is arranged as a
$T\times C_{\ell}\times N_{\ell}$ tensor, spatially averaged while preserving
time, and projected onto a fixed rank-$4$ scaffold.

This interface is kept unchanged across the two backbones. The architectural
transfer experiment therefore changes the internal firing and residual
organization of the predictor while preserving the PAFS extraction,
construction, and correction protocol.

\paragraph{Backbone papers and implementations.}
The official resources are:

\begin{itemize}
    \item \textbf{Spikformer}~\cite{zhou2023spikformer}:
    official implementation of the ICLR 2023 model,
    available at
    \url{https://github.com/ZK-Zhou/spikformer}.

    \item \textbf{Spikingformer}~\cite{zhou2026spikingformer}:
    official implementation accompanying the cited paper,
    available at
    \url{https://github.com/TheBrainLab/Spikingformer}.
\end{itemize}

\paragraph{Temporal corruption operators.}
The main paper describes the temporal distortions using event-time
terminology at a high level. In the evaluated implementation,
timestamp jitter and speed change are applied after event-to-frame
conversion to the fixed $T=16$ event-count representation.
Timestamp jitter stochastically displaces discrete frame-bin contents,
whereas speed change resamples the temporal frame axis and restores
the original sequence length by zero-padding. The exact operators are
specified below.

\paragraph{Window Drop.}
For a drop fraction $\rho=0.3$, the number of removed bins is
\[
w
=
\min\!\left(
T,
\max\!\left(1,\operatorname{round}(T\rho)\right)
\right).
\]
With $T=16$, this gives $w=5$. For each sample $n$, the starting bin is drawn
independently as
\[
a_n
\sim
\operatorname{DiscreteUniform}\{0,\ldots,T-w\},
\]
so that $a_n\in\{0,\ldots,11\}$ in the reported setting. The bins in
$[a_n,a_n+w)$ are set to zero, while all remaining bins are unchanged.
This corruption therefore removes one contiguous interval of temporal
evidence.

\paragraph{Jitter.}
For each sample $n$ and source frame bin $t$, an independent integer offset is
sampled as
\[
\epsilon_{n,t}
\sim
\operatorname{DiscreteUniform}\{-2,-1,0,1,2\}.
\]
The destination index is
\[
t'_{n,t}
=
\operatorname{clip}(t+\epsilon_{n,t},0,T-1).
\]
The complete content of source bin $t$ is accumulated at destination bin
$t'_{n,t}$. When multiple source bins map to the same destination, their values
are summed. Offsets that would leave the valid range are accumulated at the
first or final bin; bins are neither discarded nor wrapped around. Because
this operator acts on already-binned count frames, it does not perform
event-level timestamp reordering.

\paragraph{Speed.}
For speed factor $s$, the temporal length is first changed to
\[
T'
=
\max\!\left(1,\operatorname{round}(T/s)\right).
\]
The reported setting uses the acceleration convention $s=2.0$, giving
$T'=8$. The complete temporal axis is linearly interpolated from 16 to 8 bins
with PyTorch interpolation using $\mathrm{align\_corners}=\mathrm{false}$. The result is then center-padded with four zero bins on each side to restore length 16. Thus, the compressed signal occupies output bins 4 through 11. Event counts are not explicitly renormalized after interpolation, and the operator does not separately remove events outside an observation window because the entire frame sequence is compressed before padding.

\paragraph{Controlled data usage and comparison protocol.}
Training the Adapter and correction head, estimating normalization statistics,
and constructing the firing scaffolds use only samples from the training split.
Hyperparameters, checkpoints, and the correction scale are selected exclusively
on the validation split, while the test split is reserved for final reporting.
Within each dataset--backbone setting, all methods use the same
pretrained checkpoint, data splits, frame conversion, corruption
definitions, validation and test realizations, and matched random
seeds. All model and scale selection is performed exclusively on
the validation split using the predefined criterion for the
corresponding method or training stage.

\paragraph{Metrics aligned with the evaluation claims.}
The metrics distinguish preservation of the original predictor from robustness
under temporal distribution shift. Clean accuracy measures whether calibration
degrades the uncorrupted operating condition. Accuracy under each individual
corruption exposes condition-specific behavior and prevents gains on one
temporal shift from masking failures on another.

Let $p_0$ denote the clean condition,
$\mathcal P_{\rm rob}$ the set containing Window Drop, Jitter, and Speed, and
$\mathcal P=\{p_0\}\cup\mathcal P_{\rm rob}$. For $S$ matched runs, the
cross-run mean accuracy under condition $p$ is
\begin{equation}
\overline{\mathrm{Acc}}(p)
=
\frac{1}{S}
\sum_{s=1}^{S}
\mathrm{Acc}_{s}(p).
\label{eq:supp_mean_accuracy}
\end{equation}
The aggregate metrics are
\begin{equation}
\begin{aligned}
\mathrm{Avg}
&=
\frac{1}{|\mathcal P|}
\sum_{p\in\mathcal P}
\overline{\mathrm{Acc}}(p),\\
\mathrm{CorrAvg}
&=
\frac{1}{|\mathcal P_{\rm rob}|}
\sum_{p\in\mathcal P_{\rm rob}}
\overline{\mathrm{Acc}}(p),\\
\mathrm{Worst}
&=
\min_{p\in\mathcal P_{\rm rob}}
\overline{\mathrm{Acc}}(p).
\end{aligned}
\label{eq:supp_metrics}
\end{equation}

Avg. summarizes performance under a uniformly weighted mixture of clean and
corrupted conditions. Corr.~Avg. excludes the typically high clean accuracy so
that robustness gains are not diluted by the uncorrupted condition. Worst
reports the lowest cross-run mean among the three corruptions and therefore
captures performance under the most challenging evaluated temporal shift. The
primary robustness criteria are Corr.~Avg. and Worst, while clean accuracy is
reported as a preservation constraint.

All accuracies are reported in percent as mean $\pm$ standard deviation over
matched runs. For Worst, we first identify the corruption with the lowest
cross-run mean accuracy and then report the standard deviation of that same
corruption, ensuring that its mean and variation refer to one common operating
condition. Differences between methods are computed from matched runs, and all
reported robustness gains are absolute percentage-point differences.

\FloatBarrier

\section{Appendix C: Reproducibility Settings and Hyperparameters}
\label{sec:supp_reproducibility}
\begin{table}[t]
\centering
\footnotesize
\setlength{\tabcolsep}{3.0pt}
\renewcommand{\arraystretch}{1.48}
\begin{tabularx}{\columnwidth}{
@{}
>{\raggedright\arraybackslash}p{0.20\columnwidth}
>{\raggedright\arraybackslash}p{0.45\columnwidth}
>{\raggedright\arraybackslash}X
@{}
}
\toprule
Control
& Matched components
& Changed factor \\
\midrule

LoRA
& Frozen predictor, data protocol, and validation selection
& Parameter-efficient weight update \\

Plain MLP
& Frozen predictor and approximately matched Adapter budget
& Output-space residual MLP without an Adapter anchor \\

Logit-only
& Adapter anchor, correction budget, optimization, and selection
& Uses only $\phi(z)$; no firing features \\

PCA / Random
& Anchor, LIF sources, rank, correction head, and training
& Fixed firing-basis construction rule \\

Scalar rules
& Candidate coordinates, retained budget, correction head, and training
& Channel-ranking criterion only \\
\bottomrule
\end{tabularx}
\caption{
Matched conditions and isolated factors for the baseline and control
comparisons.
}
\label{tab:supp_control_configurations}
\end{table}

\paragraph{Reproducibility protocol.}
All quantities used to construct PAFS are estimated exclusively from the
training split. The validation split is used for checkpoint, hyperparameter,
and correction-scale selection, while the test split is reserved for final
reporting. The three stages are explicitly decoupled: the pretrained backbone
and original classifier remain frozen throughout, the Adapter is frozen before
scaffold construction, and only the correction head is optimized after the
scaffold has been fixed. All comparison methods use matched data splits, corruption
realizations, validation data, and random seeds. Checkpoints and
deployment scales are selected using the predefined validation-only
criterion for the corresponding method or training stage.
\paragraph{Calibration sampling and data use.}
PAFS requests a training-only calibration pool with cap
$N_{\mathrm{req}}$. Let $N_{\mathrm{available}}$ denote the number
of examples available in the corresponding checkpoint training split.
The realized pool size is
\[
N_{\mathrm{pool}}
=
\min\!\left(
N_{\mathrm{req}},
N_{\mathrm{available}}
\right).
\]
The default request cap is $N_{\mathrm{req}}=512$. In the default
DailyAction-DVS setting, all requested examples are available, so
$N_{\mathrm{pool}}=512$.

For each matched model seed, the realized pool is independently and
deterministically partitioned using a seed derived from that model seed
and the scaffold-fitting fraction $f_{\mathrm{fit}}=0.75$.
The resulting subsets are disjoint:
$f_{\rm fit}=0.75$. 
\[
\mathcal D_{\mathrm{pool}}
=
\mathcal D_{\mathrm{fit}}
\mathbin{\dot\cup}
\mathcal D_{\mathrm{rest}}.
\]
Under the default setting, $N_{\mathrm{fit}}=384$ unique training examples are used for normalization-statistic estimation and scaffold construction. The remaining $N_{\mathrm{rest}}=128$ examples are excluded from Stage~2 scaffold estimation. This partition is generated independently for each matched model seed and does not define an additional validation split.

Because Adapter and correction-head optimization use the complete
training split $\mathcal D_{\mathrm{tr}}$, samples from both
$\mathcal D_{\mathrm{fit}}$ and $\mathcal D_{\mathrm{rest}}$ may
appear as ordinary training examples in Stages~1 and~3.
The $\mathcal D_{\mathrm{rest}}$ subset is not used to estimate
scaffold statistics or bases, select a checkpoint or deployment
scale, or report final test performance.

To align notation with the main paper, we use $\mathcal D_{\mathrm{cal}}\equiv\mathcal D_{\mathrm{fit}}$ and $N_{\mathrm{cal}}=N_{\mathrm{fit}}=384$. The default value $N_{\rm req}=512$ denotes the calibration-pool request cap, whereas $N_{\rm cal}=N_{\rm fit}=384$ is the number of examples used to estimate normalization statistics and scaffold bases.

\paragraph{Adapter-anchor optimization.}
During Stage~1, only the Adapter parameters are optimized; the pretrained
backbone and original classifier remain fixed. Both trainable stages use
AdamW~\citep{loshchilov2019adamw}; only the optimized parameter set differs. Each mini-batch is drawn from
one of the four training domains---Clean, Window Drop, Jitter, or Speed---as
specified by the multi-domain objective in Eq.~(1) of the main paper. This
protocol exposes the Adapter to all evaluated operating conditions without
mixing different temporal distributions within one mini-batch.

The Adapter checkpoint is selected by mean validation accuracy over the clean
condition and the three temporal corruptions. This criterion favors an anchor
that performs consistently across operating conditions rather than specializing
in either clean inputs or a single corruption. Once selected, the Adapter is
frozen and denoted by $\eta^\star$. It then defines both the firing trajectories
and prediction residuals used in scaffold construction.

\paragraph{Scaffold construction.}
Stage~2 contains no gradient-based optimization. The frozen backbone,
classifier, and Adapter anchor are used to extract the selected late-layer LIF
trajectories from the scaffold-fitting subset. Channel--time normalization,
probability-residual computation, error-aware weighting, domain-balanced
association, and SVD-based basis extraction follow Eqs.~(2)--(12) of the main
paper. The resulting normalization statistics and rank-$r$ scaffold bases are
stored as fixed buffers and receive no gradients during correction training.

Freezing the Adapter before this stage ensures that the firing trajectories and
prediction residuals remain defined relative to one common anchor predictor.
Allowing the Adapter to change afterward would alter the representation
coordinate system and create a moving target for the correction head.
\paragraph{Correction-head optimization and scale selection.}
During Stage~3, only the linear correction head is optimized. The backbone,
original classifier, Adapter anchor, normalization statistics, and scaffold
bases remain fixed. Correction training follows the same
one-domain-per-mini-batch protocol used for Adapter training.

The correction weight and bias are initialized to zero, so the initial
residual satisfies $\Delta z_\omega=0$ and training begins exactly from the
frozen Adapter anchor. The correction head is trained with a fixed unit
residual scale,
\[
\alpha_{\rm train}=1,
\qquad
z^{\rm train}(x)=z(x)+\Delta z_\omega(x).
\]
Writing $x^p=\mathcal T_p(x)$, the resulting trained correction head is
\begin{equation}
\begin{aligned}
\omega^\star
=
\arg\min_\omega\;
\mathbb E_{\substack{(x,y)\sim\mathcal D_{\rm tr}\\
p\sim{\rm Unif}(\mathcal P)}}
\Big[
&\operatorname{CE}
\big(
z(x^p)+\Delta z_\omega(x^p),y
\big)
\\
&+
\lambda_\Delta
\|
\Delta z_\omega(x^p)
\|_2^2
\Big].
\end{aligned}
\label{eq:supp_correction_training}
\end{equation}

The deployment scale is selected only after $\omega^\star$ has been fixed.
The nonzero correction scales summarized in the main paper are
\[
\mathcal A_{+}=\{0.25,0.5,0.75,1.0\}.
\]
The validation procedure additionally retains $\alpha=0$ as an
anchor-preserving fallback, giving
\[
\mathcal A=\{0\}\cup\mathcal A_{+}.
\]
For $\alpha=0$, the corrected predictor exactly recovers the frozen
Adapter anchor:
\[
z+\alpha\Delta z_{\omega^\star}=z.
\]
We first retain the scales that satisfy the anchor-relative clean-performance
guard:
\begin{equation}
\begin{gathered}
\mathcal A_\epsilon
=\Big\{\alpha\in\mathcal A:\\
\operatorname{Acc}^{\rm val}_{\rm clean}
\big(z+\alpha\Delta z_{\omega^\star}\big)
\ge
\operatorname{Acc}^{\rm val}_{\rm clean}(z)-\epsilon\Big\},\\
\epsilon=0.005 .
\end{gathered}
\label{eq:supp_feasible_scales}
\end{equation}
Accuracy is represented on the $[0,1]$ scale, so
$\epsilon=0.005$ permits a maximum clean-validation reduction of
0.5 percentage points.

Among the feasible candidates, we select
\begin{equation}
\hat{\alpha}
=
\arg\max_{\alpha\in\mathcal A_\epsilon}
\operatorname{Acc}^{\rm val}_{\rm robust}
\!\left(
z+\alpha\Delta z_{\omega^\star}
\right),
\label{eq:supp_scale_selection}
\end{equation}
where $\operatorname{Acc}^{\rm val}_{\rm robust}$ is the mean validation
accuracy over Window Drop, Jitter, and Speed. The deployed predictor is
therefore
\begin{equation}
z^{\rm corr}(x)
=
z(x)
+
\hat{\alpha}\Delta z_{\omega^\star}(x).
\label{eq:supp_deployed_correction}
\end{equation}
Accordingly, Eq.~(18) of the main paper is instantiated during
correction-head optimization with the fixed unit scale
$\alpha_{\rm train}=1$. The candidate deployment scales in
$\mathcal A$ are not used to fit $\omega$; they are evaluated only after
training has produced and frozen $\omega^\star$. Test-time prediction then
uses Eq.~\eqref{eq:supp_deployed_correction} with the validation-selected
scale $\hat{\alpha}$. Both $\omega^\star$ and $\hat{\alpha}$ remain fixed
during test evaluation.

\begin{table}[t]
\centering
\footnotesize
\setlength{\tabcolsep}{3.5pt}
\renewcommand{\arraystretch}{1.22}
\begin{tabularx}{\columnwidth}{
@{}
>{\raggedright\arraybackslash}p{0.42\columnwidth}
>{\raggedright\arraybackslash}X
@{}
}
\toprule
Setting & Value \\
\midrule
Input representation
& $T=16$ event-count frames \\

Spatial resolution
& $128\times128$ \\

Adapter bottleneck
& $b=12$ \\

Adapter dropout
& $0.05$ \\

Optimizer
& AdamW \\

Learning rate
& $5\times10^{-4}$ \\

Batch size
& 128 \\

Maximum Adapter epochs
& 160 \\

Maximum correction-head epochs
& 80 \\

Adapter checkpoint criterion
& Four-condition mean validation accuracy \\

Correction-head training scale
& $\alpha_{\rm train}=1$ \\

Correction-head initialization
& Zero weight and zero bias \\

Selected LIF sources
& Attention \textrm{proj\_lif} and MLP \textrm{fc1\_lif} \\

Selected layers
& Two late LIF layers \\

Scaffold rank
& $r=4$ per selected layer \\

Error weight
& $\gamma=4$ \\

Correction penalty
& $\lambda_{\Delta}=10^{-3}$ \\

Nonzero deployment scales &
$\{0.25,0.5,0.75,1.0\}$ \\

Anchor-preserving fallback &
$\alpha=0$ \\

Deployment-scale criterion &
Validation Corr.\ Avg.\ subject to the clean-performance guard \\

Clean-performance guard
& $\epsilon=0.005$ ($0.5$ percentage points) \\

Calibration request cap
& $N_{\rm req}=512$ \\

Scaffold-fitting fraction
& $f_{\rm fit}=0.75$ \\

Default realized split
& 512 pool $=$ 384 scaffold-fitting $+$ 128
scaffold-excluded samples \\

Matched random seeds
& 7, 11, and 42 \\
\bottomrule
\end{tabularx}
\caption{
Default PAFS implementation, optimization, and model-selection settings.
The calibration request cap applies to unique training samples and may be
truncated by training-split availability.
}
\label{tab:supp_hyperparameters}
\end{table}

\paragraph{Corruption resampling and evaluation seeds.}
Temporal corruption parameters are generated deterministically from the
matched model seed and the current execution stage. During Adapter, LoRA, and
Plain-MLP training, one of the four domains is selected for each mini-batch,
and stochastic corruption parameters are regenerated when the epoch or batch
step changes. Thus, training exposes each method to repeatedly resampled
temporal perturbations rather than to one fixed corrupted training set.

During correction-head training, the clean, Window Drop, Jitter, and Speed
domains are traversed according to the recorded epoch--step schedule, and the
stochastic parameters of the selected corruption are regenerated for each
mini-batch.

For validation, test evaluation, and deployment-scale selection, the
domain-specific random generator is recreated from the same deterministic seed
for every evaluation call. Consequently, each method, anchor/corrected
predictor pair, and candidate residual scale is evaluated on the same
corruption realization within a matched model seed. Validation and test
corruptions therefore remain fixed across checkpoints and scale candidates,
whereas training corruptions are resampled over optimization steps.

\paragraph{Controlled data usage and comparison protocol.}
Training the Adapter and correction head, estimating normalization statistics,
and constructing the firing scaffolds use only samples from the training split.
Checkpoints, hyperparameters, and deployment scales are selected exclusively
on the validation split using the predefined criterion for the corresponding
method or training stage, while the test split is reserved for final reporting.
Within each dataset--backbone setting, all methods use the same pretrained
checkpoint, data splits, frame conversion, corruption definitions, validation
and test realizations, and matched model seeds. These controls isolate the
effect of the adaptation or correction mechanism from differences in data
access, corruption sampling, or evaluation conditions.

\paragraph{Parameter-efficient and output-space controls.}
As summarized in Table~\ref{tab:supp_control_configurations}, the Adapter,
LoRA, and Plain MLP baselines evaluate alternative lightweight adaptation or
output-space correction mechanisms. The more tightly matched Logit-only MLP
isolates whether anchor output statistics alone are sufficient for residual
correction.

The Adapter anchor is the predictor obtained after Stage~1 and contains neither
scaffold projection nor residual correction. It is the direct reference for
measuring the contribution of Stages~2 and~3.

\paragraph{Baseline implementations.}
LoRA~\citep{hu2022lora} is applied directly to the frozen Full predictor.
It uses rank~8, scaling parameter~16, dropout~0.05, and adapts the query,
key, value, attention-output, and two MLP projections in the final encoder
block. The original projections remain frozen, and the second low-rank factor
is initialized to zero.

Plain MLP on Full attaches a two-layer residual MLP with ReLU and
dropout~0.05 to the logit descriptor $\phi(z)$ of the original frozen
checkpoint. Its output projection is initialized to zero, so the initial
predictor is identical to the Full checkpoint. The hidden widths are selected as the nearest parameter match to the 6,924-parameter Adapter: 173, 28, and 203 for DailyAction-DVS,
DVS-GC, and CIFAR10-DVS, respectively.

The Logit-only MLP is instead attached to the same frozen Adapter anchor as
PAFS and receives only $\phi(z)$, without firing hooks or scaffold features.
Its two-layer residual head is parameter-matched to the corresponding PAFS
correction head, and its output projection is initialized to zero. Thus,
correction training begins exactly from the Adapter anchor.

The PCA and random-scaffold controls use the same frozen Adapter anchor,
selected LIF sources, retained rank, correction head, and optimization
protocol as PAFS. PCA retains the leading principal-component directions of
the standardized firing trajectories. The random scaffold uses a fixed
Gaussian basis orthogonalized by reduced QR decomposition. Scalar channel
controls instead retain individual firing channels through a one-hot basis;
only the channel-ranking criterion changes.

\paragraph{Baseline optimization.}
Adapter, LoRA, and Plain MLP on Full use AdamW with learning rate
$5\times10^{-4}$, zero weight decay, batch size~128, global-norm gradient
clipping at~1.0, and at most 160 epochs. Validation is performed every five
epochs, with early-stopping patience of six validation events. These methods
use CUDA automatic mixed precision during training and FP32 evaluation.

PAFS and the Logit-only, PCA, random-scaffold, and scalar-selection controls
use AdamW with the same learning rate and zero weight decay, batch size~128,
global-norm gradient clipping at~1.0, and at most 80 correction epochs.
Correction training is performed in FP32 with the residual scale fixed to
$\alpha_{\rm train}=1$. The trained correction head is subsequently frozen,
and the deployment scale is selected only on the validation split according
to the protocol above.
\paragraph{Firing-space controls.}

The firing-space controls in
Table~\ref{tab:supp_control_configurations} preserve the frozen Adapter anchor,
correction capacity, and optimization protocol while changing how the retained
firing representation is constructed. The PCA and random scaffold controls preserve the same Adapter anchor, firing
sources, rank, correction head, and optimization procedure as PAFS. PCA uses
the leading principal-component directions of standardized firing trajectories,
whereas the random control uses a fixed random orthonormal basis of the same
shape. They test whether activation-dominant or arbitrary low-dimensional
firing representations can replace residual-aligned scaffold construction.

The scalar channel controls preserve the same candidate features,
retained-feature budget, correction head, and optimization protocol while
changing only the ranking criterion. Random ranking, activation frequency,
activation variance, and temporal variance test generic firing salience.
Perturbation sensitivity favors channels that change between clean and
corrupted inputs, while residual relevance explicitly scores association with
the anchor prediction residual. These controls distinguish individual-channel
importance from multivariate residual-aligned firing structure.

\paragraph{Software, hardware, and random seeds.}
Experiments use Ubuntu 22.04.5, Python 3.12.3, PyTorch 2.8.0 with
CUDA 12.8 and cuDNN 9.10.2. The server contains four NVIDIA GeForce
RTX 5090 GPUs. Each training or evaluation run uses one GPU; independent
seeds and configurations may run in parallel, but no individual run uses
distributed multi-GPU optimization. The main dependencies include
SpikingJelly 0.0.0.0.8, Tonic 1.6.0, CuPy-CUDA12x 13.6.0,
timm 1.0.27, NumPy 1.26.4, SciPy 1.14.1, and scikit-learn 1.5.1.
Unless otherwise stated, results use matched model seeds 7, 11, and 42.

\FloatBarrier

\section{Appendix D: Additional Experimental Results}
\label{sec:supp_experiments}

\paragraph{Overview.}
The following analyses test three practical and mechanistic
implications of the PAFS hypothesis. First, we examine whether
useful residual-aligned scaffolds can be estimated from a modest
training-only calibration pool. Second, we test whether the robustness gain extends beyond matched
correction capacity when scaffold-projected internal firing
coordinates are provided. Third, we test whether generic
or independently ranked firing features can replace the proposed
multivariate residual-aligned scaffold.

\begin{figure}[!hbtp]
\centering
\includegraphics[width=\columnwidth]
{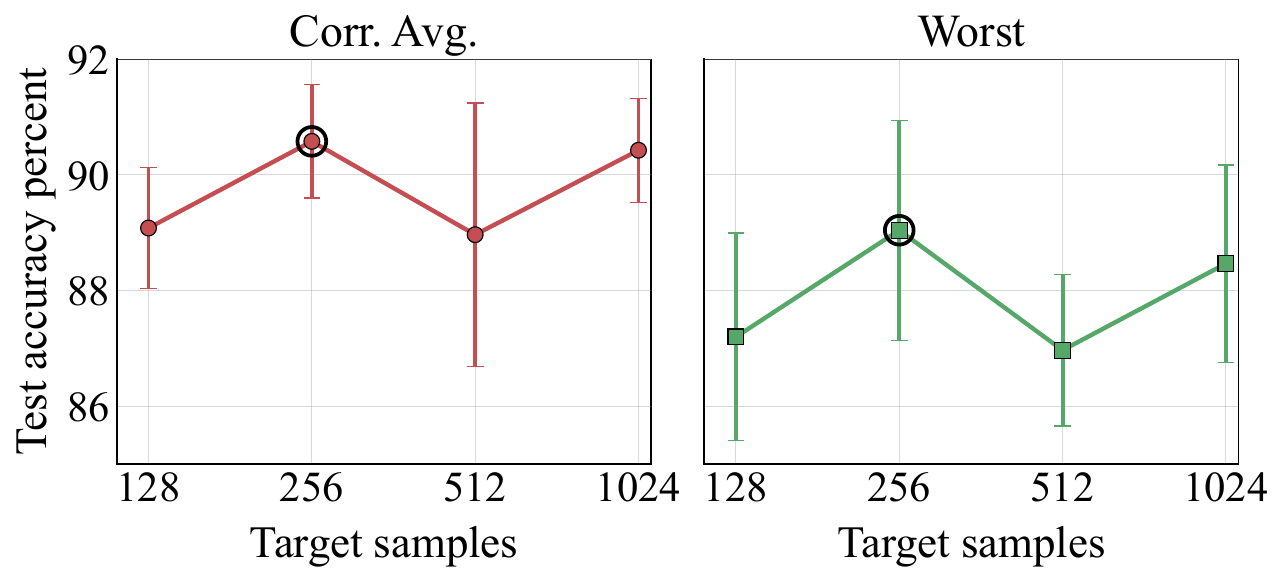}
\caption{
Sensitivity to the training-only calibration request cap on DailyAction-DVS
with Spikformer. Requested caps of 128, 256, 512, and 1,024 yield realized
pools of 128, 256, 512, and 929 unique samples, respectively. The fixed
0.75 fitting fraction gives scaffold-fitting subsets of 96, 192, 384, and
697 samples. Each point is the mean over three matched seeds, with error bars
denoting $\pm1$ standard deviation. Corr.~Avg. averages Window Drop, Jitter,
and Speed; Worst reports the corruption with the lowest cross-run mean.
}
\label{fig:calibration_budget}
\end{figure}

\paragraph{Estimating residual-aligned scaffolds from limited calibration data.}
A practical consequence of the proposed low-dimensional firing
representation is that its residual-aligned directions should be
recoverable from a modest training-only calibration pool.
Figure~\ref{fig:calibration_budget} therefore varies the number
of unique training samples available for scaffold construction.

\renewcommand{\dbltopfraction}{0.95}
\renewcommand{\dblfloatpagefraction}{0.95}
\setlength{\dbltextfloatsep}{6pt}
\begin{table*}[ht]
\centering
{\small
\setlength{\tabcolsep}{3.2pt}
\renewcommand{\arraystretch}{1.0}

\begin{tabular*}{\textwidth}
{@{\extracolsep{\fill}}lrrrrrrr@{}}
\toprule
& \multicolumn{3}{c}{Deployment overhead $\downarrow$}
& \multicolumn{2}{c}{Measured inference cost $\downarrow$}
& \multicolumn{2}{c}{Robustness gain vs.\ Adapter $\uparrow$} \\
\cmidrule(lr){2-4}
\cmidrule(lr){5-6}
\cmidrule(lr){7-8}

Method
& \shortstack{Learned params\\($K$ / \% frozen)}
& \shortstack{FP32 storage\\(KiB)}
& \shortstack{Added MACs\\($M$/sample)}
& \shortstack{Latency\\(ms/sample)}
& \shortstack{Energy\\(mJ/sample)}
& \shortstack{$\Delta$ Corr.\ Avg.\\(pp)}
& \shortstack{$\Delta$ Worst\\(pp)}
\\
\midrule

Adapter
& $6.924$ / $0.270\%$
& $27.0$
& $6.291$
& $5.989\pm0.048$
& $819.1\pm11.1$
& $0.00$
& $0.00$
\\

LoRA
& $36.864$ / $1.435\%$
& $144.0$
& \shortstack[r]{$37.749$\\$(+31.458)$}
& \shortstack[r]{$6.299\pm0.160$\\$(+5.2\%)$}
& \shortstack[r]{$854.9\pm12.7$\\$(+4.4\%)$}
& $-1.12$
& $+0.12$
\\

Logit-only MLP
& $8.796$ / $0.342\%$
& $34.4$
& \shortstack[r]{$6.293$\\$(+0.002)$}
& \shortstack[r]{$6.318\pm0.065$\\$(+5.5\%)$}
& \shortstack[r]{$853.3\pm3.3$\\$(+4.2\%)$}
& $-0.83$
& $+0.24$
\\

\addlinespace[2pt]

\textbf{PAFS (ours)}
& $8.796$ / $0.342\%$
& $106.4$
& \shortstack[r]{$6.326$\\$(+0.035)$}
& \shortstack[r]{$6.626\pm0.011$\\$(+10.6\%)$}
& \shortstack[r]{$880.1\pm7.4$\\$(+7.5\%)$}
& $\mathbf{+1.84}$
& $\mathbf{+3.23}$
\\

\bottomrule
\end{tabular*}
}

\caption{
Efficiency--robustness trade-off on DailyAction-DVS with Spikingformer.
All methods share the same frozen $2.569$M-parameter predictor.
Learned parameters and FP32 storage denote method-specific additions;
PAFS storage additionally includes $72.0$~KiB fixed scaffold and
normalization buffers.
Added MACs are measured relative to the frozen predictor.
Latency and energy are mean $\pm$ standard deviation over three matched
seeds. Parentheses indicate changes from Adapter.
Robustness gains are absolute percentage-point improvements.
Here, $K=10^3$, $M=10^6$, and KiB denotes $2^{10}$ bytes.
}

\label{tab:efficiency_supp}
\end{table*}

\begin{table*}[!htbp]
\centering
{\small
\renewcommand{\arraystretch}{0.90}
\setlength{\tabcolsep}{2.2pt}
\begin{tabular*}{\textwidth}
{@{\extracolsep{\fill}}lrrrrrr@{}}
\toprule
Method / scoring rule
& Clean
& Window Drop
& Jitter
& Speed
& Corr. Avg.
& Worst
\\
\midrule
Adapter
& 97.35$\pm$0.53
& 86.62$\pm$1.22
& 90.43$\pm$2.50
& 86.85$\pm$3.65
& 87.97$\pm$1.57
& 86.62$\pm$1.22
\\
PAFS (Ours)
& 97.46$\pm$0.53
& 87.20$\pm$1.51
& 90.77$\pm$2.80
& \textbf{88.93$\pm$3.34}
& \textbf{88.97$\pm$2.28}
& \textbf{87.20$\pm$1.51}
\\
\midrule
Random ranking
& 97.18$\pm$0.61
& 85.74$\pm$1.68
& 89.88$\pm$2.17
& 85.92$\pm$3.08
& 87.18$\pm$1.46
& 85.74$\pm$1.68
\\
Activation frequency
& 97.52$\pm$0.48
& 86.48$\pm$1.45
& 90.61$\pm$1.92
& 86.34$\pm$2.84
& 87.81$\pm$1.30
& 86.34$\pm$2.84
\\
Activation variance
& \textbf{97.58$\pm$0.46}
& 86.71$\pm$1.56
& 89.92$\pm$2.06
& 86.10$\pm$2.73
& 87.58$\pm$1.28
& 86.10$\pm$2.73
\\
Temporal variance
& 97.31$\pm$0.55
& \textbf{87.36$\pm$1.41}
& 89.85$\pm$1.80
& 86.20$\pm$2.69
& 87.80$\pm$1.21
& 86.20$\pm$2.69
\\
Perturbation sensitivity
& 97.28$\pm$0.52
& 86.95$\pm$1.29
& 90.55$\pm$2.11
& 87.62$\pm$2.91
& 88.37$\pm$1.38
& 86.95$\pm$1.29
\\
Residual relevance
& 97.41$\pm$0.47
& 87.08$\pm$1.33
& \textbf{90.94$\pm$1.71}
& 87.85$\pm$2.51
& 88.62$\pm$1.15
& 87.08$\pm$1.33
\\
\bottomrule
\end{tabular*}
}
\caption{
Controlled comparison of PAFS and scalar channel-selection rules on
DailyAction-DVS with Spikformer under the default rank-4-per-layer
protocol. PAFS and all scalar controls use the same retained-feature
budget and the same 1,872-parameter correction head. Only the rule used
to construct the retained firing representation changes. Results are mean test accuracy (\%) $\pm$ standard deviation over three matched model seeds. Corr.~Avg. averages Window Drop, Jitter, and Speed, while Worst denotes the corruption with the lowest cross-run mean accuracy.
}
\label{tab:selection_rule_supp}
\end{table*}
With $f_{\rm fit}=0.75$ held fixed, the four realized pools
produce scaffold-fitting subsets of 96, 192, 384, and 697 samples and scaffold-excluded remainder subsets of 32, 64, 128, and 232 samples. The firing sources,
scaffold rank, correction head, corruption protocol, and model-selection
procedure remain unchanged.

Increasing the request cap from 128 to 256, corresponding to an
increase from 96 to 192 scaffold-fitting samples, improves both
Corr.~Avg. and Worst. The trend is non-monotonic thereafter: the
predefined 512-cap configuration remains competitive, while the
1,024 request---realized as 929 samples---does not consistently
outperform the smaller configurations.

The non-monotonic behavior indicates that robustness does not
increase mechanically with scaffold-fitting set size. Within this
fixed-rank study, useful residual-aligned directions can already be
estimated from a modest training-only subset. This experiment
therefore supports scaffold estimability under a limited calibration
budget, but does not independently establish intrinsic low rank or
monotonic sample efficiency. The default request cap $N_{\rm req}=512$ was fixed before this
sensitivity analysis and was not selected from the reported test
curves.

\paragraph{Efficiency--robustness trade-off.}
Table~\ref{tab:efficiency_supp} reports a representative deployment profile on
DailyAction-DVS with Spikingformer. All methods operate on the same frozen
2.569M-parameter predictor and are measured under identical profiling
conditions.

PAFS and the matched Logit-only MLP introduce the same number of learned parameters,
isolating the value of scaffold-projected firing information from correction
capacity. Relative to Adapter, the Logit-only MLP reduces Corr. Avg. by 0.83 percentage points while improving Worst by 0.24 points. PAFS instead improves Corr. Avg. and Worst by 1.84 and 3.23 points, respectively.

This improvement adds 0.035M MACs over Adapter, while measured latency and
gross GPU board energy increase by 10.6\% and 7.5\%. The main runtime cost
comes from collecting two late-layer firing trajectories and projecting them
onto fixed rank-4 bases; no additional backbone pass is required. The matched
comparison therefore indicates that the robustness gain is associated with
internal firing coordinates rather than output-space correction capacity alone.

\paragraph{Residual alignment and multivariate firing structure.}
Table~\ref{tab:selection_rule_supp} evaluates increasingly targeted
alternatives to the proposed scaffold, ranging from generic firing
salience and perturbation sensitivity to independently ranked
residual-relevant channels. All variants preserve the same Adapter
anchor, firing sources, retained-feature budget, correction head, and
optimization protocol; only the construction of the retained firing
representation changes.

Random ranking and generic activation statistics do not consistently
outperform the Adapter anchor, indicating that firing magnitude or
variability alone is insufficient for identifying correction-relevant
internal responses. Perturbation sensitivity performs more strongly,
suggesting that channels affected by temporal shifts are more informative
than generally active channels. Residual relevance is the strongest scalar
control, reaching 88.62 Corr.~Avg. and 87.08 Worst. This result supports
the interpretation that alignment with the Adapter anchor's remaining
prediction errors is more informative than generic firing salience or
perturbation sensitivity alone.

PAFS achieves the strongest overall robustness, with 88.97 Corr.~Avg.
and 87.20 Worst, outperforming every scalar ranking rule under the same
retained-feature budget and correction protocol. It also obtains the best
Speed accuracy, exceeding residual relevance by 1.08 percentage points
under global temporal rescaling. Although individual scalar controls are
competitive under specific corruptions, none matches PAFS on both aggregate
robustness criteria. Together, these comparisons indicate that residual
alignment accounts for most of the observed robustness improvement in this
controlled setting, while the multivariate functional-SVD scaffold captures
complementary joint firing structure that is not recovered by independently
ranking channels.
\FloatBarrier

\FloatBarrier
\clearpage
\renewcommand{\refname}{Supplementary References}
\putbib[paper/references,paper/supplementary_references]
\end{bibunit}
\end{document}